\documentclass[10pt,twocolumn,letterpaper]{article}

\usepackage{wacv}
\usepackage{times}
\usepackage{epsfig}
\usepackage{graphicx}
\usepackage{amsmath}
\usepackage{amssymb}
\usepackage{booktabs}
\usepackage{subfigure}
\usepackage{caption}

\usepackage{algorithm}
\usepackage{algpseudocode}
\usepackage{amsmath}
\usepackage{amssymb}
\usepackage{mathtools}
\usepackage{amsthm}
\usepackage[hang,flushmargin]{footmisc}

\def\wacvPaperID{102} 
\wacvfinalcopy 

\newcommand*\samethanks[1][\value{footnote}]{\footnotemark[#1]}
\makeatletter
\def\blfootnote{\gdef\@thefnmark{}\@footnotetext}
\makeatother

\ifwacvfinal
\usepackage[breaklinks=true,bookmarks=false]{hyperref}
\else
\usepackage[pagebackref=true,breaklinks=true,colorlinks,bookmarks=false]{hyperref}
\fi

\begin{document}

\title{Controllable Style Transfer via Test-time Training of Implicit Neural Representation}

\author{Sunwoo Kim\thanks{Equal contribution}
\qquad Youngjo Min\samethanks
\qquad Younghun Jung\samethanks
\qquad Seungryong Kim\thanks{Corresponding author}\\
Korea University, South Korea\\
{\tt\small \{sw-kim, 1320harry, raidria, seungryong\_kim\}@korea.ac.kr}
}

\twocolumn[
{
\renewcommand\twocolumn[1][]{#1}%
\maketitle
\begin{center}
    \centering
    \captionsetup{type=figure}
    \includegraphics[width=1.0\textwidth,height=6cm]{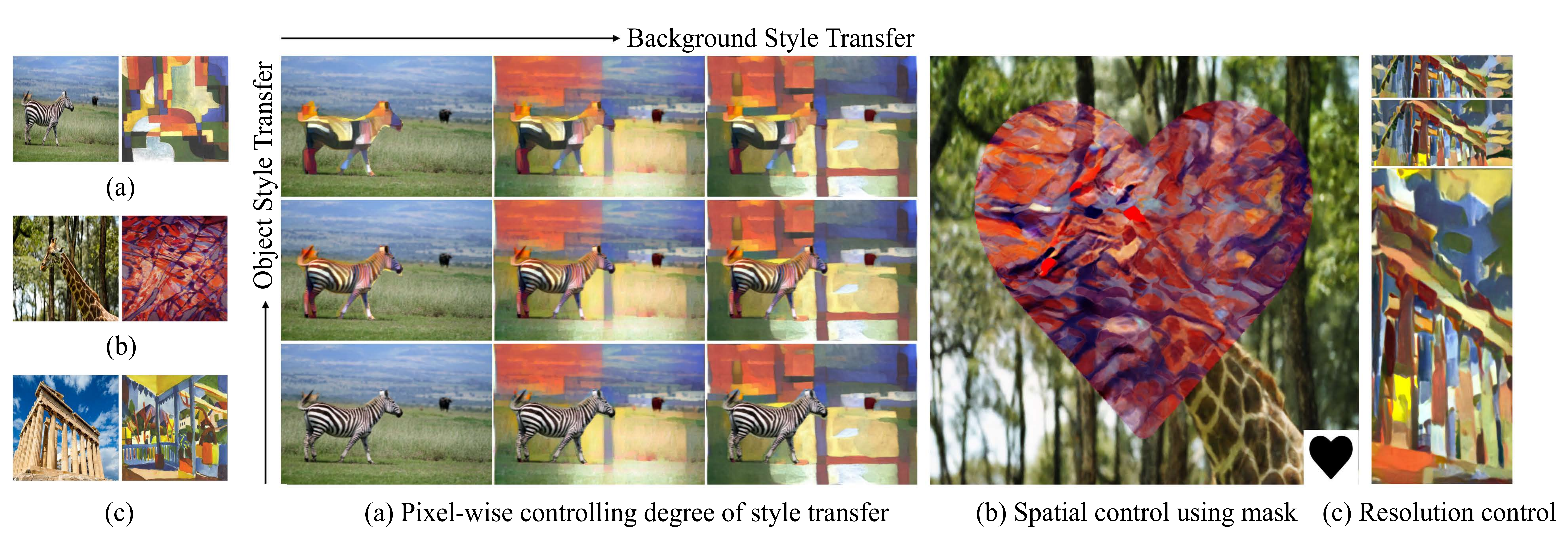}
    \vspace{-20pt}
    \captionof{figure}{\textbf{Results of our controllable style transfer framework.} If our INR-based network is optimized at test time once, it can perform a flexible pixel-wise stylization of given content and style images without re-optimization at inference. Our method can change the degree of style applied to the background along the horizontal axis or adjust the degree of style for the foreground (zebra) along the vertical axis in (a). (b) demonstrates that our method can selectively stylize a specific region of the image by using a mask in the bottom right corner. In both (a) and (b), we can observe that only the target regions are stylized with the desired degrees of style without affecting neighboring pixels. (c) shows the unconstrained manipulation of the image resolution and aspect ratio conducted by our model.}
    \label{fig:teaser}
\end{center}%
}
]
\thispagestyle{empty}

\begin{abstract}
\blfootnote{
$*$Equal contribution \\  
$\dagger$ Corresponding author
}
We propose a controllable style transfer framework based on Implicit Neural Representation that pixel-wisely controls the stylized output via test-time training. Unlike traditional image optimization methods that often suffer from unstable convergence and learning-based methods that require intensive training and have limited generalization ability, we present a model optimization framework that optimizes the neural networks during test-time with explicit loss functions for style transfer. After being test-time trained once, thanks to the flexibility of the INR-based model, our framework can precisely control the stylized images in a pixel-wise manner and freely adjust image resolution without further optimization or training. We demonstrate several applications. Our code is available at ~\url{https://KU-CVLAB.github.io/INR-st/}.


\end{abstract}

\section{Introduction}
A style transfer task aims to render an image with style features of an artistic image while preserving the structure of a content image~\cite{gatys2016image}. Among several early works, Gatys et al.~\cite{gatys2015texture,gatys2016image} first introduced convolutional neural network (CNN)-based style transfer algorithm that jointly minimizes content and style losses. After this work, subsequent image optimization-based models~\cite{li2016combining,gu2018arbitrary, li2017demystifying, risser2017stable} have explored improvements, but they have a common problem that they often fall into a bad local minimum due to unstable training.

\begin{figure*}[t!] 
\centering
    \renewcommand{\thesubfigure}{}
    \subfigure[(a) Optimization-based methods]
	{\includegraphics[width=0.33\linewidth]{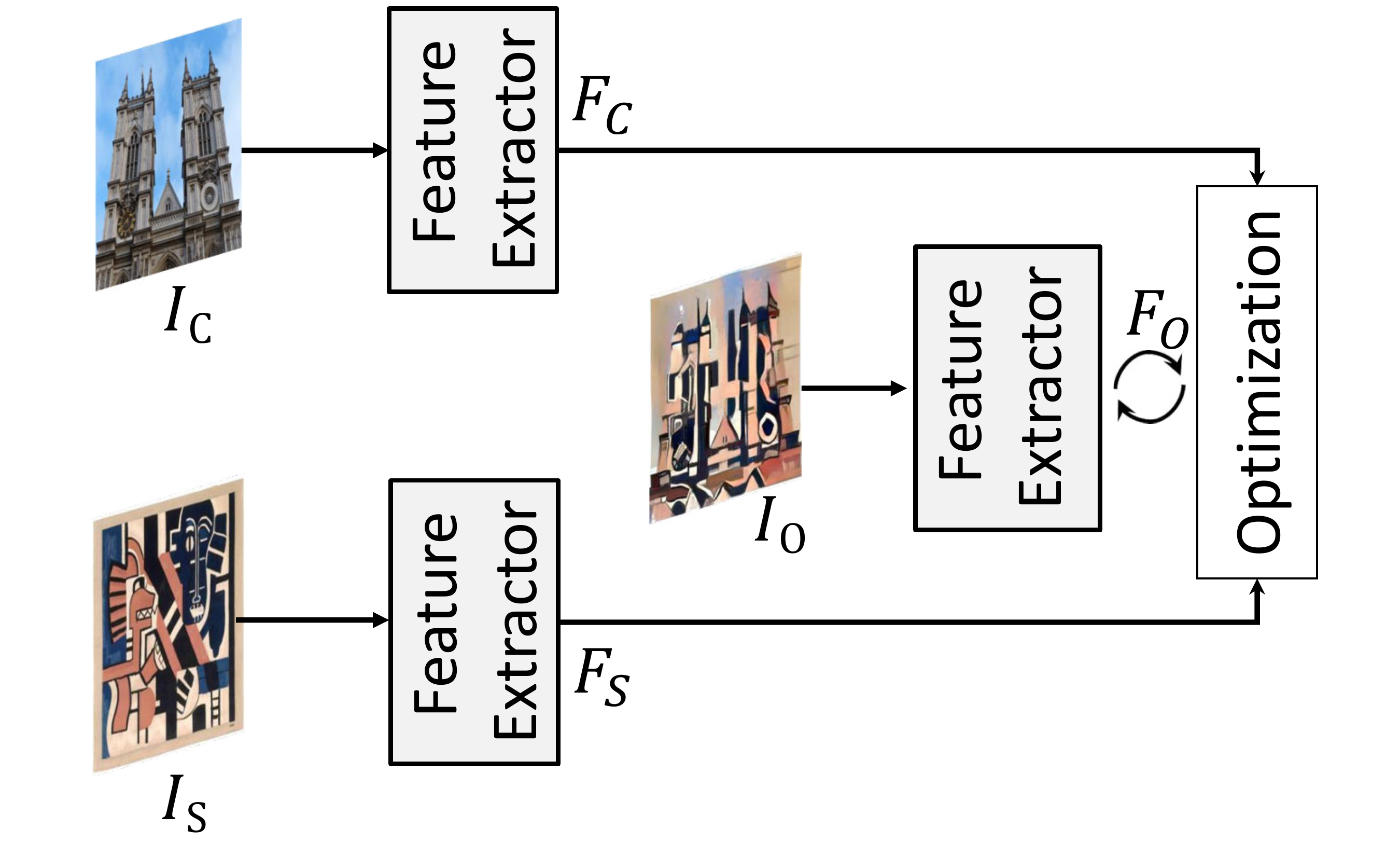}}\hfill
	\subfigure[(b) Learning-based methods]
	{\includegraphics[width=0.33\linewidth]{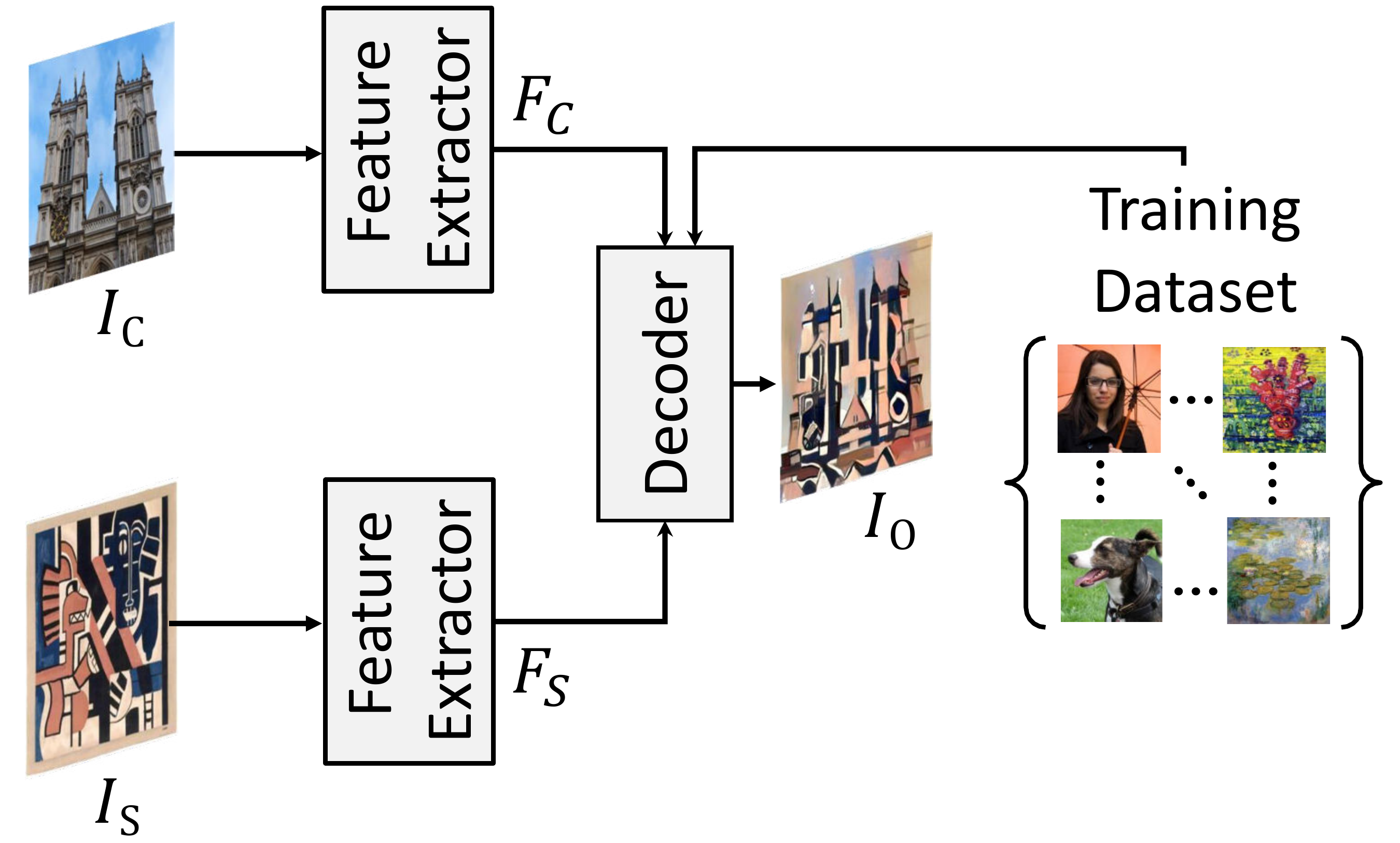}}\hfill
	\subfigure[(c) Ours]
	{\includegraphics[width=0.33\linewidth]{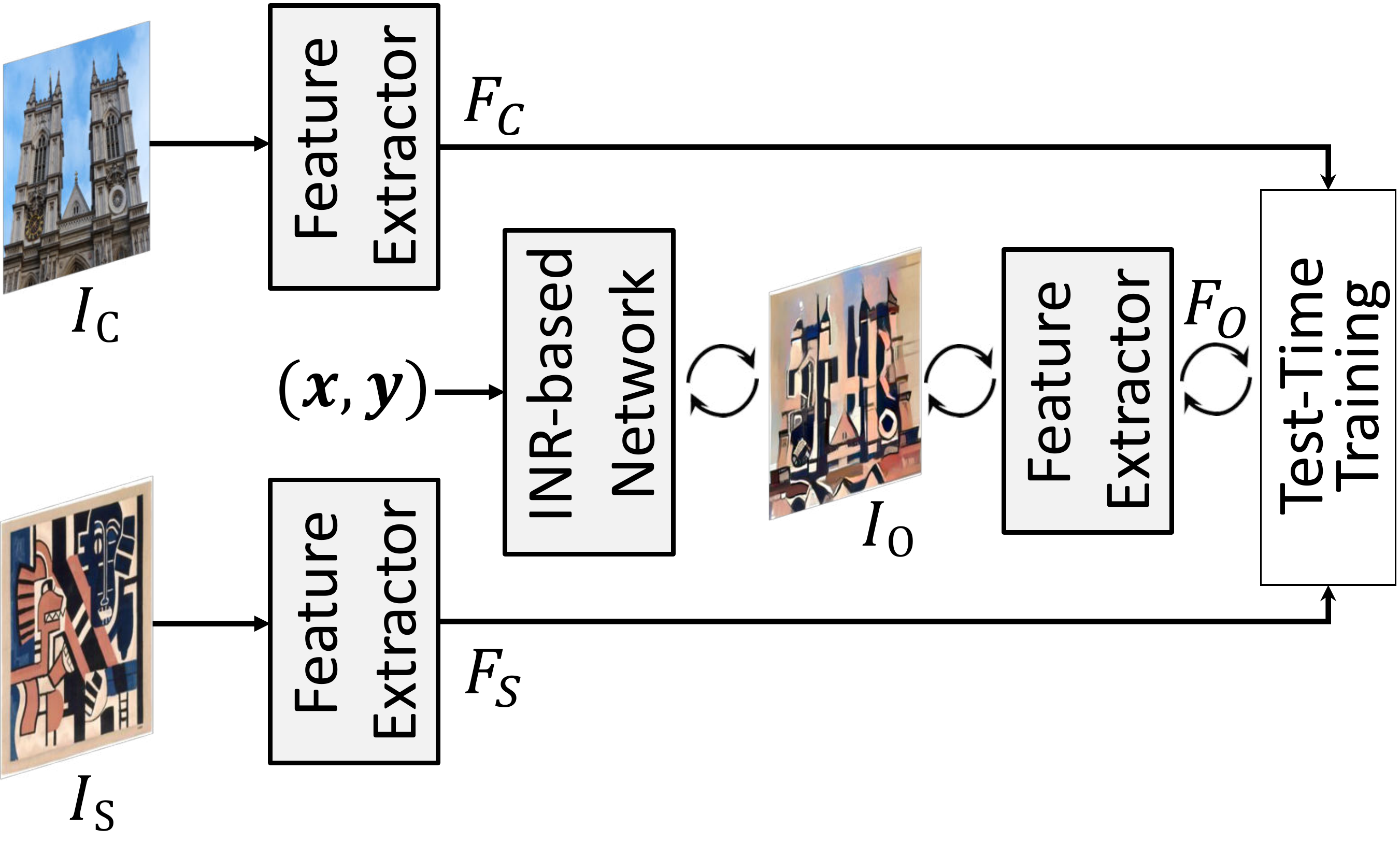}}\hfill\\
	\vspace{-10pt}
	\caption{\textbf{Intuition of our model.} (a) image optimization-based methods~\cite{gatys2016image,li2016combining} that optimize an output image itself at test-time with explicit content and style losses, which often generate artifacts and produce limited performance due to the lack of image prior, (b) learning-based methods that require intensive offline training on large-scale training data and use pre-trained, fixed networks at test time~\cite{huang2017arbitrary,li2017universal,li2018closed,gu2018arbitrary}, and (c) our model that learns the image prior with coordinate-based neural networks at test-time and makes controllable and continuous representation.}\label{fig:intuition}
	\vspace{-10pt}    
\end{figure*}

Another major direction is to train a CNN-based neural network to perform style transfer, often called a learning-based approach. Pioneering works in this direction were restricted to a set of predefined styles~\cite{johnson2016perceptual,ulyanov2016texture,ulyanov2016instance}, and following works mitigated this constraint by leveraging adaptive instance normalization (AdaIN) layer~\cite{huang2017arbitrary}, style-swap method~\cite{chen2016fast}, whitening and coloring processes (WCT)~\cite{li2017universal}, attention mechanism~\cite{park2019arbitrary,liu2021adaattn}, and contrastive learning~\cite{chen2021artistic, zhang2022cast}. However, since these learning-based algorithms require a large-scale training dataset and lack generalization ability on unseen images, they have limited applicability.

\begin{figure}[t!] \centering
    \renewcommand{\thesubfigure}{}
    \subfigure[(a) Ours]
	{\includegraphics[width=0.325\linewidth]{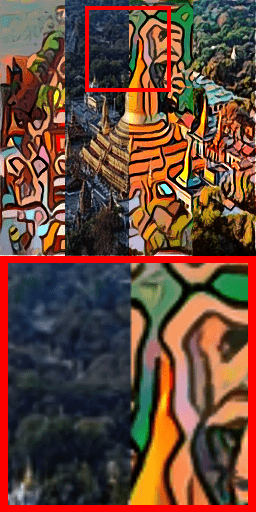}}\hfill
    \subfigure[(b) SANet ]
	{\includegraphics[width=0.325\linewidth]{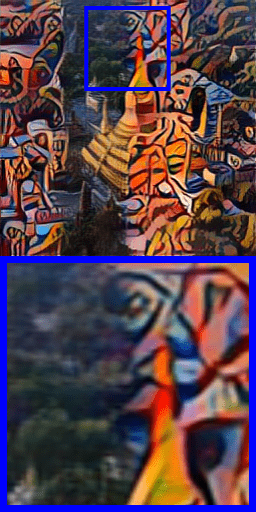}}\hfill
    \subfigure[(c) AdaAttn]
    {\includegraphics[width=0.325\linewidth]{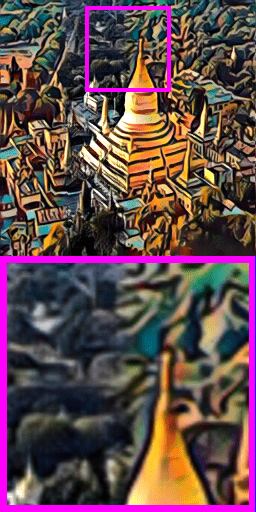}}\hfill
    
   \vspace{-10pt}
  \caption{\textbf{Pixel-wise synthesis.} Each area within an image is stylized with a different degree of $\alpha$ (from the left, 0.25, 1.0, 0.0, 0.5). Our approach can precisely transfer the desired degree of style to the target area. However, other CNN-based methods show unclear separation at the boundary where a degree of style changes as their generator performs kernel-wise operations on given image features.}
  
\label{stylesplit}
\end{figure}

\begin{figure*}[t]
    \renewcommand{\thesubfigure}{}
    \subfigure[]
	{\includegraphics[width=0.07\linewidth]{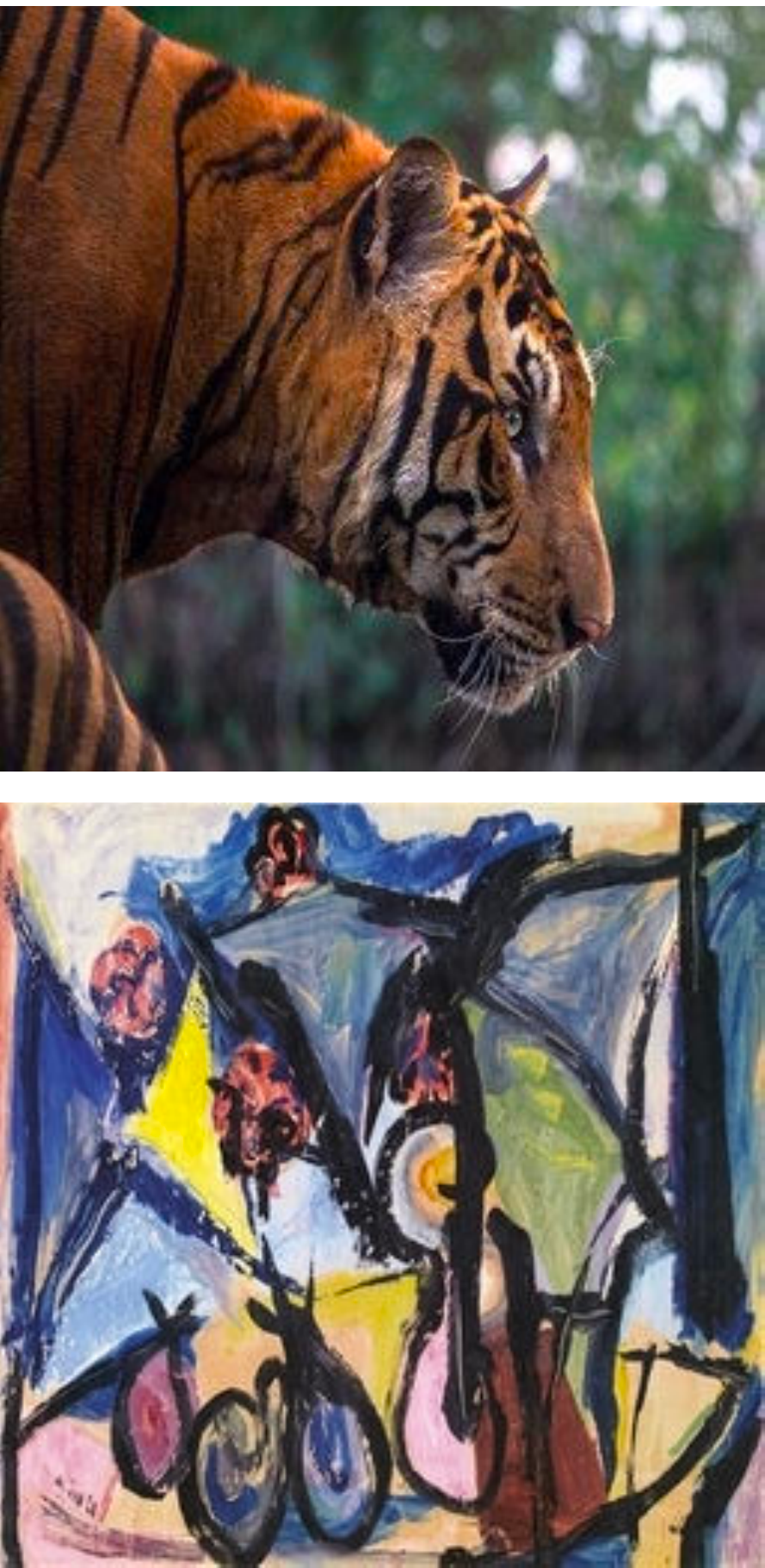}}\hfill
    \subfigure[(a) Upsampling the stylized image ]
	{\includegraphics[width=0.46\linewidth]{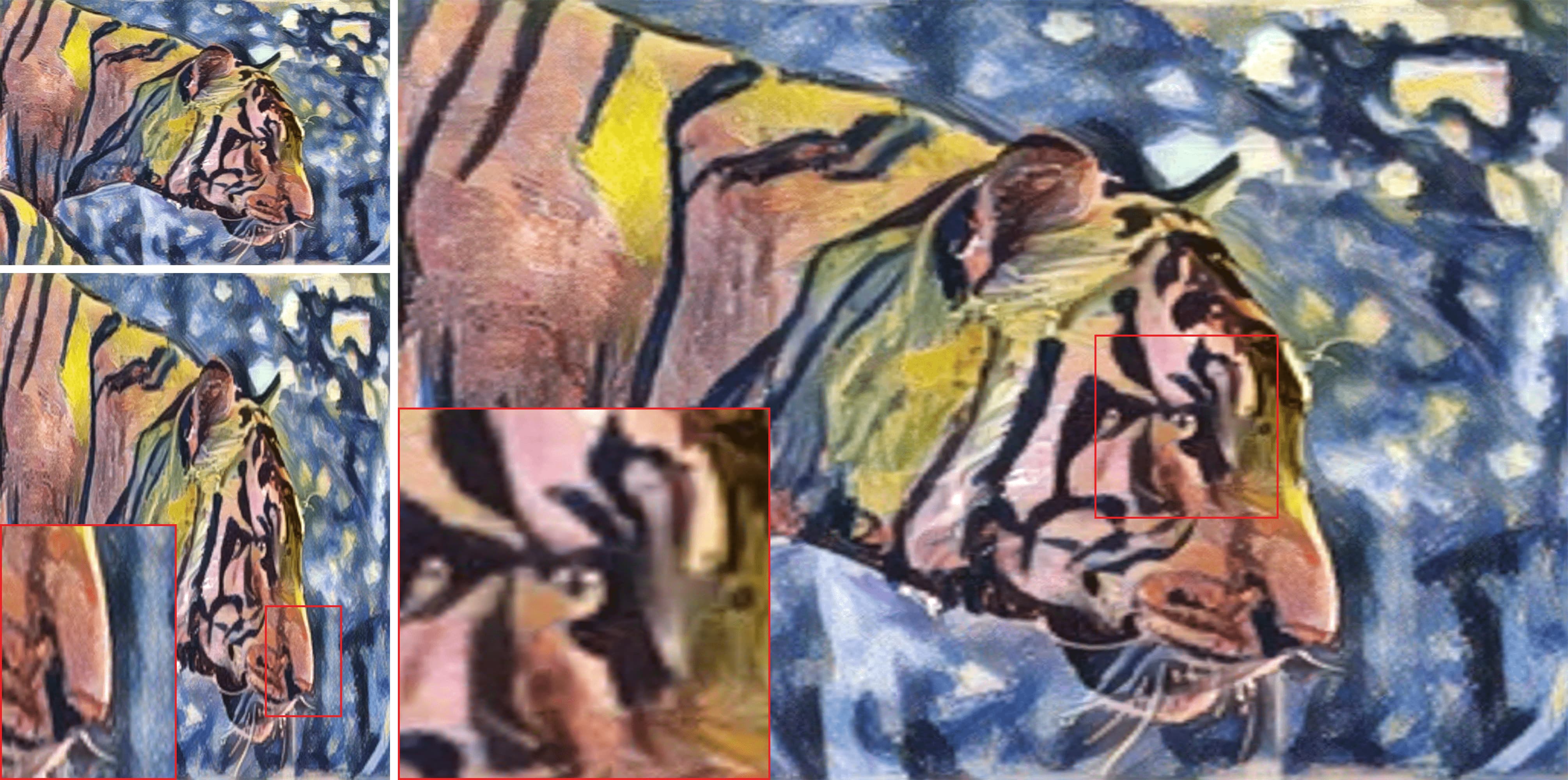}}\hfill
    \subfigure[(b) Controlling the coordinates (Ours)]
    {\includegraphics[width=0.46\linewidth]{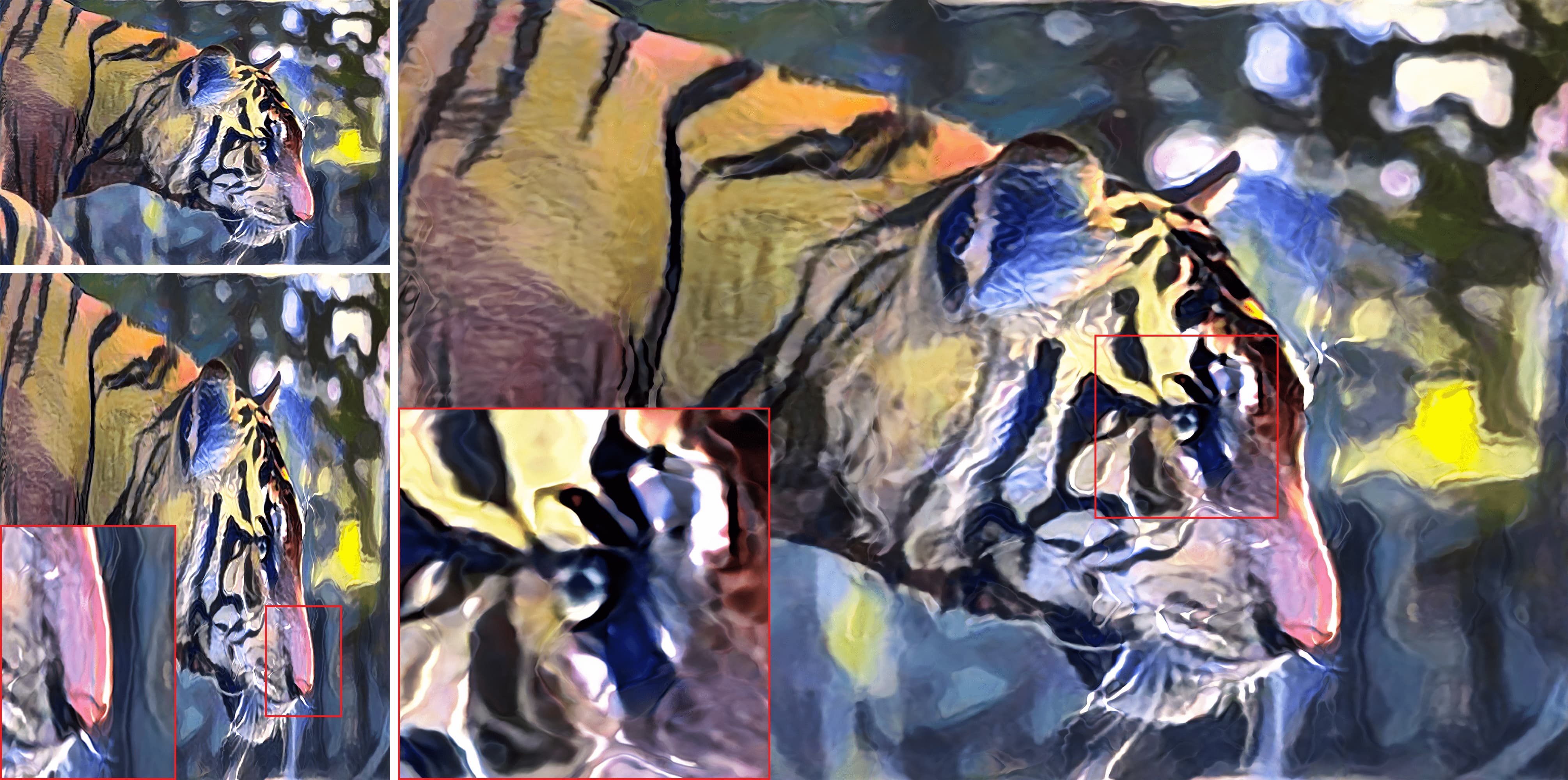}}\hfill
\vspace{-10pt}

\caption{\textbf{Comparison of high-resolution image quality.} Once our network is converged, high-fidelity images with arbitrary resolution and aspect ratio can be produced without further training. For each case, (a) and (b), the resolutions of the images are 680$\times$1,000, 2,000$\times$3,000, and 1,300$\times$1,000 respectively, in a clockwise direction from the upper left image. All images in (a) are bilinearly upsampled from 256 $\times$ 256 images generated by \cite{kolkin2019style}. The results in (b) are high resolution images generated by rearranging the input coordinate for our network. By magnifying the images, we could observe that the images of (b) were much clearer and sharper than the images of (a).}
\label{fig:controlRes}
\end{figure*}

In addition, these traditional CNN-based style transfer frameworks fail to precisely control the degree of style transfer at the target pixel as shown in Fig.~\ref{stylesplit} and cannot change image resolution without further training or optimization as displayed in Fig.~\ref{fig:controlRes}.

Meanwhile, Implicit Neural Representation (INR), which represents a given scene with neural networks in a continuous and pixel-wise manner, has been spotlighted as a new way to represent an image~\cite{tancik2020fourier}. Motivated by the recent success of NeRF~\cite{mildenhall2020nerf}, which reconstructs 3D scenes with INR, and other following models~\cite{park2019deepsdf, atzmon2020sal,gropp2020implicit,sitzmann2019scene,jiang2020local,peng2020convolutional,chabra2020deep, mildenhall2020nerf}, some recent works leveraging 2D images ~\cite{chen2019learning,tancik2020fourier,sitzmann2020implicit,chen2021learning} began to exploit the properties of INR.


Inspired by recent accomplishments of INR, for the first time in the field of style transfer, we introduce the INR-based model and demonstrate various applications such as pixel-by-pixel stylization control and arbitrary adjustment of image resolution, which none of the existing models can perform. Additionally, we intelligently integrated our INR-based model with a test-time training scheme to effectively learn an image pair specific prior with neural networks at the test-time stage, which enabled our model to demonstrate more stable training compared to the optimization method while showing better generalizability than the learning-based approach.
\section{Related Works}
\vspace{-5pt}

\label{sec:Related Works}

\begin{figure*}[t!]
\centering
\includegraphics[width=1.0\linewidth]{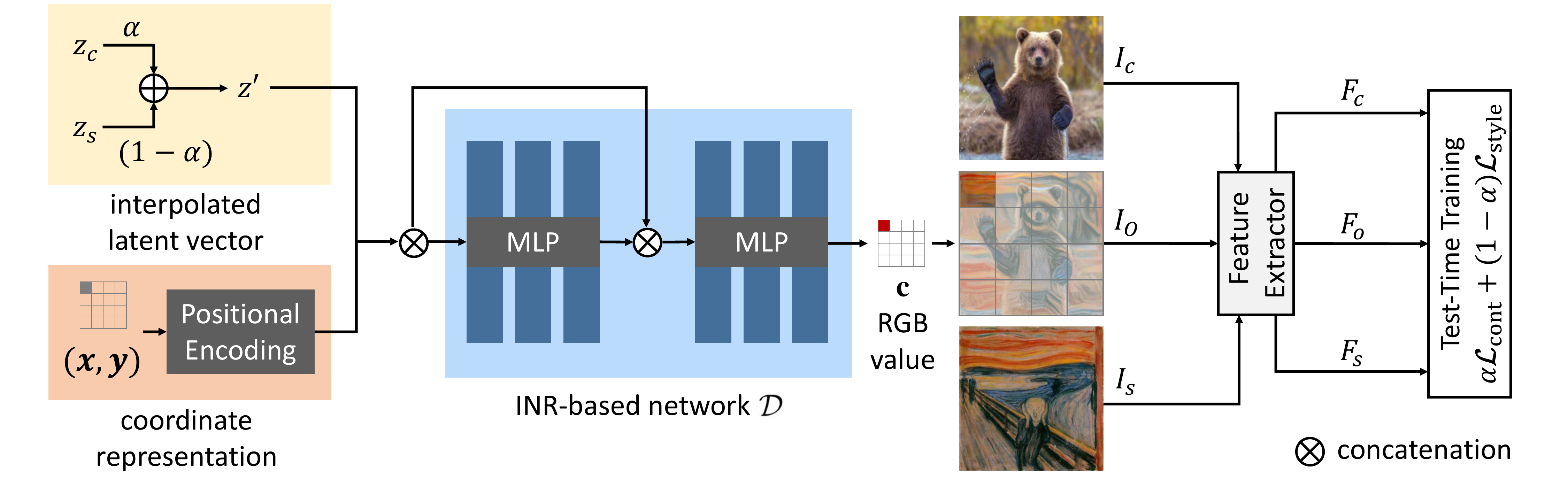} \hfill\\
    \vspace{-5pt}
	\caption{ \textbf{Network configuration of our controllable style transfer framework.}
	Our network takes two inputs; the interpolated latent vectors and the encoding of coordinate positions. We then feed-forward them to MLPs to produce RGB values. The interpolation factor $\alpha$ is renewed at every training step, which makes the network learns controllability for style transfer. 
	}
\label{fig:architecture}
\end{figure*}

\paragraph{Neural Style Transfer.} Gatys et al.~\cite{gatys2016image} initially proposed a neural style transfer algorithm that leverages features extracted from a VGG model~\cite{Simonyan15} and is optimized by jointly minimizing the content and style loss. Following this work, several optimization-based methods suggested different ways to enhance the quality of stylization~\cite{li2016combining,mechrez2018contextual}, while gradual improvement has also been made in learning-based methodology~\cite{johnson2016perceptual,ulyanov2016texture,ulyanov2016instance,dumoulin2016learned,li2017diversified,huang2017arbitrary,park2019arbitrary,liu2021adaattn,chen2021artistic,zhang2022cast}.

Recently, methods that learn the prior of images at test time by overfitting their networks to the specific pair of images have been newly proposed. SinIR~\cite{yoo2021sinir} utilizes the decoder which learns prior of a pair of images internally for various manipulation tasks such as style transfer. DTP~\cite{kim2021deep} showed enhanced performance and generalizability by explicitly learning the pair-specific prior. 

Meanwhile, several attempts~\cite{gatys2017controlling,lu2017decoder,kolkin2019style} have been made to control the output of style transfer. However, to the best of our knowledge, no previous works have attempted pixel-wise style transfer, which applies varying degrees of style to different pixels or areas of an image.
\vspace{-5pt}


\paragraph{Implicit Neural Representation.} INR can represent any given scene as a continuous function that maps coordinates to the corresponding RGB values in an image. The majority of works applying this idea are observed in 3D deep learning~\cite{sitzmann2019scene,jiang2020local,peng2020convolutional,mildenhall2020nerf,park2019deepsdf}, but recently, several works have attempted to utilize implicit neural representation for 2D image generation. Pioneering works~\cite{stanley2007compositional,mordvintsev2018differentiable} suffered from a low generation quality, but subsequent works~\cite{tancik2020fourier,sitzmann2020implicit,skorokhodov2021adversarial} enhanced the generation quality of outputs. In light of this, we introduce the INR-based style transfer model for the first time. By taking advantage of INR, we demonstrate pixel-wise, elaborate control of stylization and arbitrary adjustment of image resolution at inference.


\section{Motivation}
Neural style transfer aims at transferring the style of image $I_s$ to the content of image $I_c$ to synthesize a stylized image $I_{o}$. To achieve this, early works~\cite{gatys2016image, li2016combining, gu2018arbitrary} adopted an image optimization technique that jointly minimizes the content and style losses in the feature space of a pre-trained CNN~\cite{Simonyan15}. Specifically, let us denote deep convolutional content and style features extracted by pre-trained feature extractor $\Phi(\cdot)$ as $F_c=\Phi(I_c)$ and $F_s=\Phi(I_s)$, respectively. Then they optimize an output image $I_{o}$ via following objective function which consists of content loss $\mathcal{L}_\mathrm{cont}$ and style loss $\mathcal{L}_\mathrm{style}$ as in Fig.~\ref{fig:intuition} (a):
\begin{equation}
I_{o} =\underset{I}{\mathrm{argmin}}
\{\mathcal{L}_\mathrm{cont}(\Phi(I),F_c) + \lambda\mathcal{L}_\mathrm{style}(\Phi(I),F_s)\},
\end{equation}
where $\lambda$ is a weight parameter. Since it is often a non-convex optimization problem, most methods leverage an iterative solver, e.g., gradient descent~\cite{li2016combining,gatys2017controlling,kolkin2019style,gatys2016image,li2017universal}, and thus they benefit from an error feedback to find better stylized images. In addition, this approach enables performing style transfer with given arbitrary content and style images. However, since the output image itself is optimized without learning the stylization prior, the optimization process is often trapped in a bad local minimum and some artifacts in the output image frequently occur~\cite{chen2021dualast}. 
In addition, to control the style transfer results, e.g., controlling the resolution and getting different degree of stylization, the optimization procedure should be repeated again, which limits the applicability. 

Unlike these methods, recent learning-based models~\cite{johnson2016perceptual, ulyanov2017improved,huang2017arbitrary,li2017universal} have attempted to learn a style transfer prior within the networks by training on massive datasets. They employ a decoder module trained on a large-scale image dataset, e.g., MS-COCO~\cite{lin2014microsoft}, as illustrated in Fig.~\ref{fig:intuition} (b). These methods can be formulated as
\begin{equation}
\begin{split}
    \omega^\dagger =\underset{\omega}{\mathrm{argmin}} 
    &\sum_{n}
    \mathcal{L}_\mathrm{cont}(\Phi(\mathcal{D}(F_{c,n},F_{s,n}; \omega)),F_{c,n})\\
    &+ \mathcal{L}_\mathrm{style}(\Phi(\mathcal{D}(F_{c,n}, F_{s,n}; \omega)),F_{s,n}),
\end{split}
\end{equation}
where $F_{c,n}$ and $F_{s,n}$ are features from $n$-th image pair, and $\mathcal{D}(\cdot;\omega)$ is a feed-forward process with decoder parameters $\omega$. At test time, given $F_{c}$ and $F_{s}$, a stylization process can be formulated such that $I_{o} = \mathcal{D}(F_c,F_s;\omega^\dagger)$. These learning-based methods assume that the image priors can be learned by exploiting large training data. However, since they do not learn a prior for the specific image pair at test time, they show insufficient generalization ability when unseen images are given as input. 

Although the previous works such as SinIR~\cite{yoo2021sinir} and DTP~\cite{kim2021deep} attempted learning prior with CNN architecture at the test-time, these works were solely designed for global stylization. On the other hand, some previous works~\cite{gatys2017controlling,lu2017decoder, kolkin2019style, park2019arbitrary,liu2021adaattn} attempted to control the stylization, but they generally cannot perform precise control of style transfer results since they often employ CNN-based generator, which hinders continuous, pixel-wise modification in the output.

\section{Methodology}

\subsection{Overview and Notation}
To address the aforementioned limitations, we propose a novel Implicit Neural Representation (INR)-based style transfer framework integrated with test-time training of the model, which can learn an image pair-specific prior within the INR network during test-time training. After test-time trained once, thanks to the high capability of the INR generator for interpolating the output space, our model achieves pixel-level controllability on an output image, and thus it can perform diverse manipulations on the output without further training. 

In specific, as shown in Fig.~\ref{fig:architecture}, our network synthesizes a stylized image $I_{o} \in \mathbb{R}^{H \times W \times 3}$ with the Multi-Layer Perceptron (MLP) structure $\mathcal{D}(\cdot ; \omega)$ parameterized as $\omega$. Generation of each pixel takes pixel coordinates along with two latent vectors $z_{c}, z_{s} \in \mathbb{R}^{d_{z}}$ which correspond to content and style latent vectors, respectively. These two latent vectors are first interpolated by the sampled value $\alpha \sim U(0,1)$ at each step. Then the interpolated vector $z'(\alpha)$ is shared by all pixel coordinates $\left(x, y\right)$ during test-time training, but each pixel coordinate may have a differently interpolated vector with a different alpha ratio at inference time. In either case, all the latent vectors are grouped and denoted as $Z'(\alpha) \in \mathbb{R}^{HW \times d_{z}}$. In addition, the coordinate vectors $(x, y) \in \mathbb{R}^{HW \times 2}$ are normalized within the range of [-1, 1] and then go through the positional encoding process~\cite{mildenhall2020nerf}, which is defined as $E \in \mathbb{R}^{HW \times d_{e}}$ where $d_{e}$ is a coordinate vector dimension after the positional encoding. Finally, $Z'$ and $E$ become the input for the network after being concatenated. Given this, the model returns the RGB values $\mathbf{c} \in \left[0, 255\right]^3$ for the corresponding pixels such that aggregated into a whole image as $\mathcal{D}\left(Z'(\alpha), E;\omega \right) \mapsto I_o$.

\subsection{Test-Time Training of INR Networks}
In this section, we introduce our test-time training method for the network parameters $\omega$. The key ingredient of our approach is dynamically varying $\alpha$ at every step to allow the network to have the ability to perform controllable style transfer in terms of $\alpha$ at inference time. 

To this end, first of all, the model interpolates between two fixed latent vectors $z_{c}$ and $z_{s}$, and yields the latent vector $z'$. At every iteration of test-time training, we randomly sample a value $\alpha$ that determines how the latent vectors would be interpolated such that
\begin{equation}
z'(\alpha) = \alpha z_{c} + (1-\alpha) z_{s}, \qquad    0 \leq \alpha \leq 1.
\end{equation}
Given the latent vector $z'(\alpha)$, by using MLP structure for implicit representation as described previously, an image pair-specific prior can be captured by minimizing explicit total loss functions as:

\begin{equation}
\begin{split}
&\omega^{*} =\underset{\omega}{\mathrm{argmin}}\,\mathcal{L}_\mathrm{total}, \\
&\mathcal{L}_\mathrm{total} = \alpha\mathcal{L}_\mathrm{cont}(\Phi(\mathcal{D}(Z'(\alpha), E;\omega)),F_c) \\
& + (1-\alpha)\mathcal{L}_\mathrm{style}(\Phi(\mathcal{D}(Z'(\alpha), E;\omega)),F_s)
\end{split}
\label{eq:tto}
\end{equation}
where $\omega^{*}$ indicates optimized parameters for the input image pair, and $Z'$ is the latent vectors extended from $z'$.

After the network is trained, it can synthesize pixel-wisely manipulated image by defining $z'(\alpha)$ differently for each pixel such that $I_{o}=\mathcal{D}(Z'(\alpha), E; \omega^{*})$. Note that unlike existing optimization-based algorithms that require re-optimization, our framework does not need to repeat the optimization procedure to perform each different application of stylization control. \vspace{-5pt}

\begin{algorithm}[t]
\caption{Test-Time Training of Our Model} \label{ttoalgorithm}
\begin{algorithmic}
\Require Images $I_{c}$, $I_{s}$, encoded vectors of coordinate positions $E$, latent vectors ${z}_{c}$, ${z}_{s}$ and pretrained VGG $\Phi$
\State Initialize parameters $\omega$ of the model $\mathcal{D}$
    \While {not converged}
        \State Sample an $\alpha \sim U(0,1)$
        \State Compute $Z'(\alpha)$ with $z_{c}$, $z_{s}$ and $\alpha$
        \State Compute $I_{o} = \mathcal{D}(Z'(\alpha), E; \omega)$
		\State Compute $\mathcal{L}(I_{o},I_{c}, I_{s}; \Phi, f, \alpha)$
		\State Update $\mathbf{\omega} \leftarrow \mathbf{\omega} - \delta\nabla_{\mathbf{\omega}}\mathcal{L}$
    \EndWhile
\end{algorithmic}
\end{algorithm}

\begin{figure}[t!] \centering
    \renewcommand{\thesubfigure}{}
    \subfigure[]                {\includegraphics[width=1\linewidth]{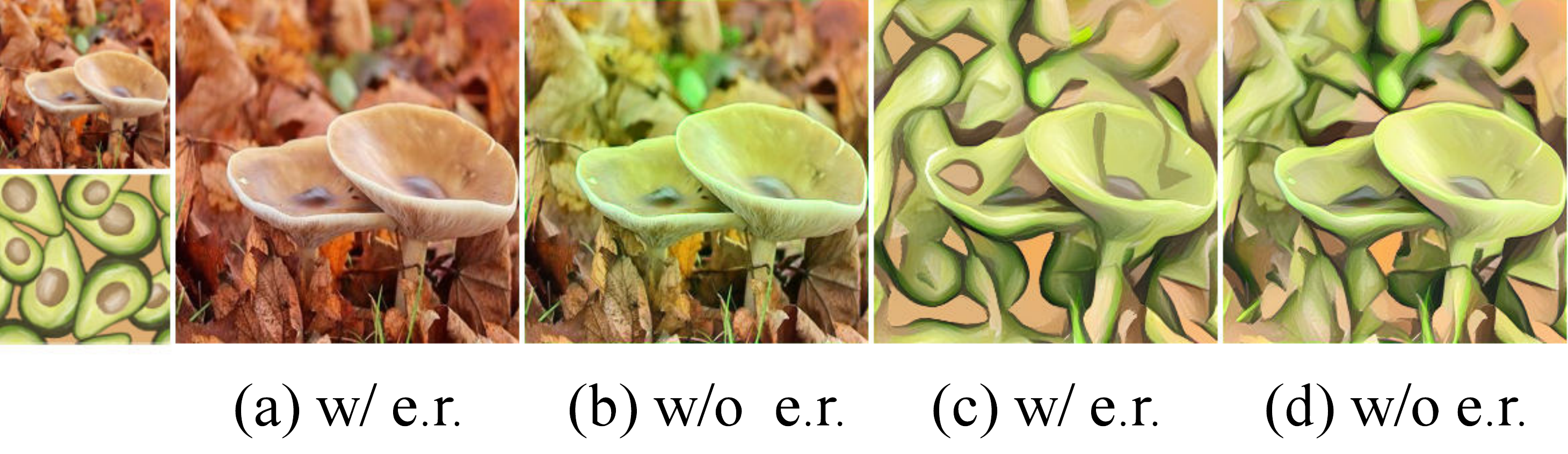}}\hfill
\vspace{-25pt}
\caption{\textbf{Effectiveness of exponentially reweighting the loss function.} The reweighting term in our loss function makes our network achieves better disentanglement between content and style during the content-style interpolation. (a), (b) are inferenced with $\alpha$ = 1 and (c), (d) are with $\alpha$ = 0.} 
\vspace{-10pt}\label{exponential}
\end{figure}

\paragraph{Exponential Reweighting.}
We experimentally found that sampling $\alpha$ in an uniformly-random manner produces a sub-optimal solution, as exemplified in Fig.~\ref{exponential}, due to the different tendency between content and style losses. Such undesirable entanglement of content and style compromises the model's controllabilty and stylization performance. Moreover, the randomly changing $\alpha$ makes the training process more unstable. For better disentanglement of content and style and relax unstable optimizing process, we devise a function to exponentially reweight the value of $x$:
\begin{equation}
    f(x) = -x\mathrm{log}(1-x^{\kappa}),
\end{equation}
where ${\kappa}$ is a hyper-parameter. It reweights $x$ as $f(x)$ to propagate the training signals exponentially. With this function, we adaptively reweight the equation~\ref{eq:tto} as follows:
\begin{equation}
\begin{split}
&\mathcal{L}_\mathrm{total} = f(\alpha)\mathcal{L}_\mathrm{cont}(\Phi(\mathcal{D}(Z'(\alpha), E;\omega)),F_c)\\
& + f(1-\alpha)\mathcal{L}_\mathrm{style}(\Phi(\mathcal{D}(Z'(\alpha), E;\omega)),F_s).\\
\end{split}
\end{equation}
This reweighting process enlarges the strength of the content property and suppresses style property when $\alpha$ is close to 1, while it amplifies the strength of style property and suppresses content property when $\alpha$ is close to 0. Additionally, increasing the value of $\kappa$ can boost the effectivenss of the reweighting process. However, assigning excessively high value to $\kappa$ could hinder the model from learning a smooth representation that adequately reflects the content and style.

\subsection{Loss Function} \label{loss}
Our framework can be incorporated with any kinds of style transfer loss functions, consisting of content and style loss function. To prove the simplicity and effectiveness of our framework, we use simple content and style losses from Gatys et al.~\cite{gatys2016image}. We employ pre-trained VGG-19~\cite{Simonyan15} as a feature extractor. Our overall loss function is the weighted summation of content loss $\mathcal{L}_\mathrm{cont}$ and style loss $\mathcal{L}_\mathrm{style}$. 

Specifically, the content loss is defined as L2 norm between the features of generate image and content image:
\begin{equation}
\mathcal{L}_\mathrm{cont} = \lambda_{{\mathrm{cont}}}\| \Phi(\mathcal{D}(Z'(\alpha),E;\omega)) - F_c \|_{2}.
\end{equation}
The style loss is defined as L2 norm between gram matrices $\mathcal{G}$ made from features of generated image and style image:
\begin{equation}
\mathcal{L}_\mathrm{style} = \lambda_{{\mathrm{style}}}\sum_{n} \| \mathcal{G}(\Phi^{n}(\mathcal{D}(Z'(\alpha),E;\omega))) - \mathcal{G}(F^n_s) \|_{2},
\end{equation}
where $n$ denotes different layer indexes of VGG encoder.

\begin{figure*} \centering
  \renewcommand{\thesubfigure}{}
  \subfigure[]  {\includegraphics[width=0.11\linewidth]{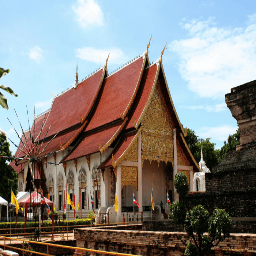}}\hfill
  \subfigure[]  {\includegraphics[width=0.11\linewidth]{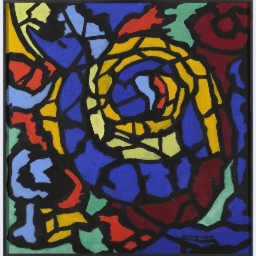}}\hfill
  \subfigure[]  {\includegraphics[width=0.11\linewidth]{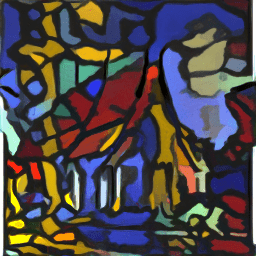}}\hfill
  \subfigure[]  {\includegraphics[width=0.11\linewidth]{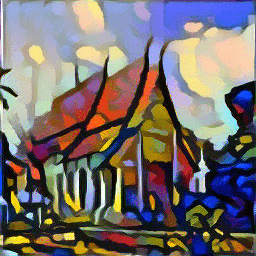}}\hfill
  \subfigure[]  {\includegraphics[width=0.11\linewidth]{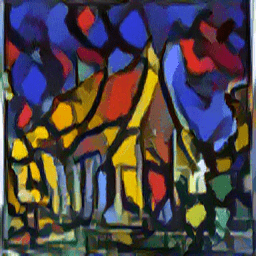}}\hfill
  \subfigure[]  {\includegraphics[width=0.11\linewidth]{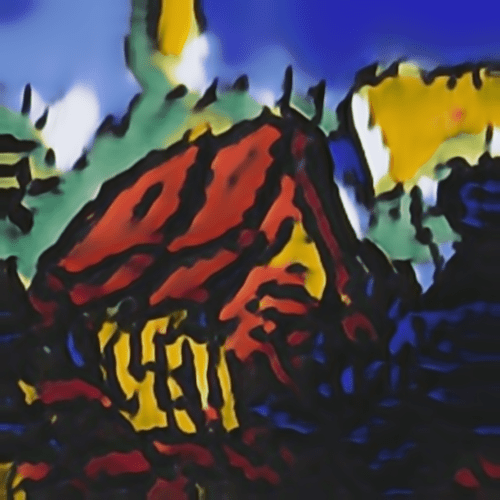}}\hfill
  \subfigure[]  {\includegraphics[width=0.11\linewidth]{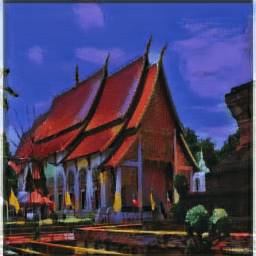}}\hfill
  \subfigure[]  {\includegraphics[width=0.11\linewidth]{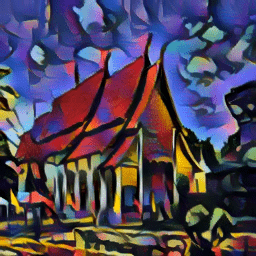}}\hfill
  \subfigure[]  {\includegraphics[width=0.11\linewidth]{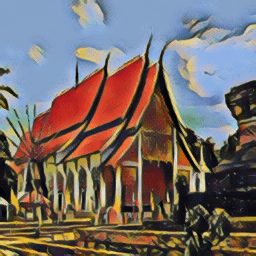}}\hfill\\
  \vspace{-22pt}

  \subfigure[]  {\includegraphics[width=0.11\linewidth]{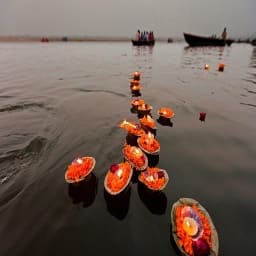}}\hfill
  \subfigure[]  {\includegraphics[width=0.11\linewidth]{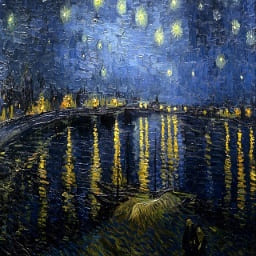}}\hfill
  \subfigure[]  {\includegraphics[width=0.11\linewidth]{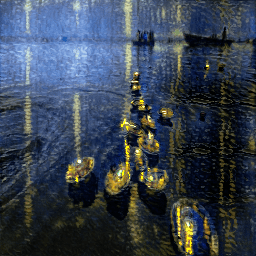}}\hfill
  \subfigure[]  {\includegraphics[width=0.11\linewidth]{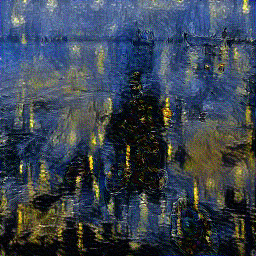}}\hfill
  \subfigure[]  {\includegraphics[width=0.11\linewidth]{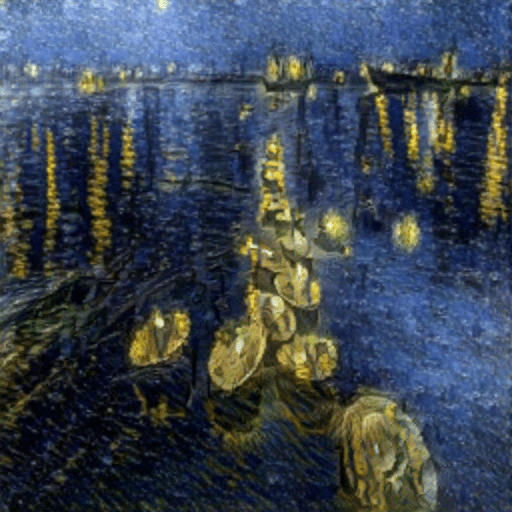}}\hfill
  \subfigure[]  {\includegraphics[width=0.11\linewidth]{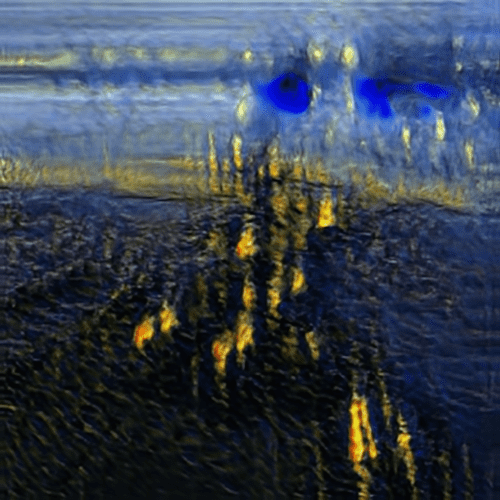}}\hfill
  \subfigure[]  {\includegraphics[width=0.11\linewidth]{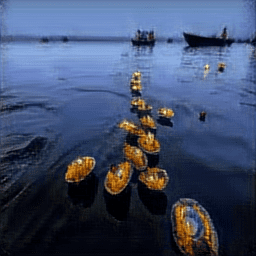}}\hfill
  \subfigure[]  {\includegraphics[width=0.11\linewidth]{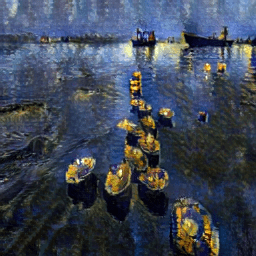}}\hfill
  \subfigure[]  {\includegraphics[width=0.11\linewidth]{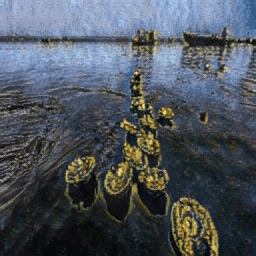}}\hfill\\
  \vspace{-22pt}
  
  \subfigure[]  {\includegraphics[width=0.11\linewidth]{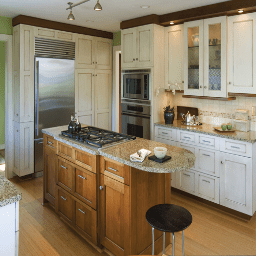}}\hfill
  \subfigure[]  {\includegraphics[width=0.11\linewidth]{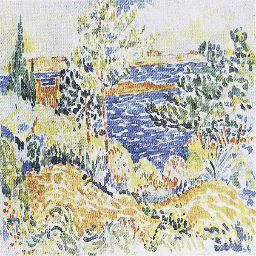}}\hfill
  \subfigure[]  {\includegraphics[width=0.11\linewidth]{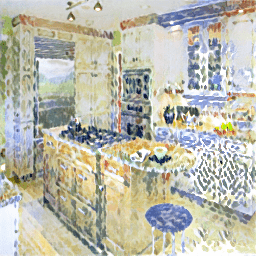}}\hfill
  \subfigure[]  {\includegraphics[width=0.11\linewidth]{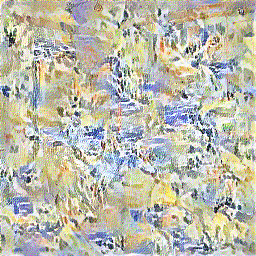}}\hfill
  \subfigure[]  {\includegraphics[width=0.11\linewidth]{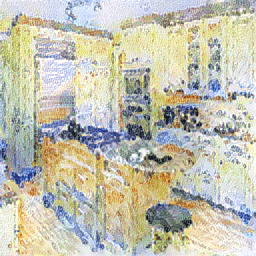}}\hfill
  \subfigure[]  {\includegraphics[width=0.11\linewidth]{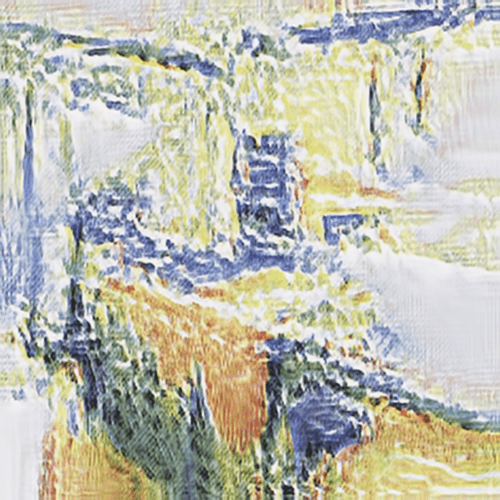}}\hfill
  \subfigure[]  {\includegraphics[width=0.11\linewidth]{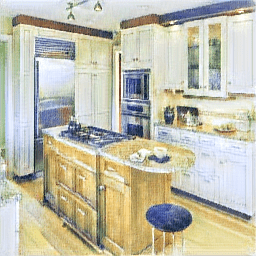}}\hfill
  \subfigure[]  {\includegraphics[width=0.11\linewidth]{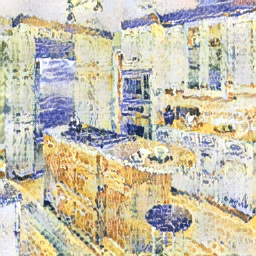}}\hfill
  \subfigure[]  {\includegraphics[width=0.11\linewidth]{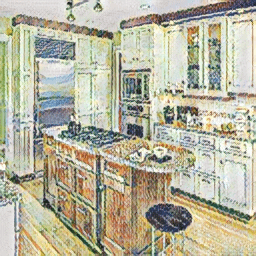}}\hfill\\ 
  
  \vspace{-22pt}
  
  \subfigure[(a) Content]
  {\includegraphics[width=0.11\linewidth]{latex/leveraging/content/bear.jpg}}\hfill
  \subfigure[(b) Style]
  {\includegraphics[width=0.11\linewidth]{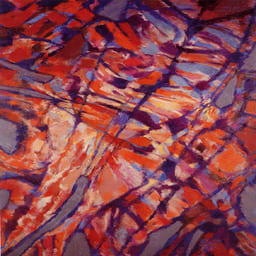}}\hfill
  \subfigure[(c) Ours]
  {\includegraphics[width=0.11\linewidth]{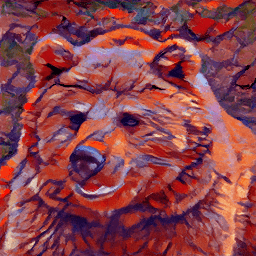}}\hfill
  \subfigure[(d) {\scriptsize Gatys et al.}]
  {\includegraphics[width=0.11\linewidth]{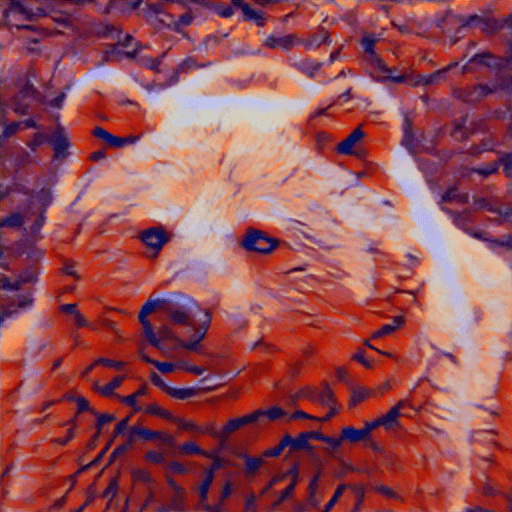}}\hfill
  \subfigure[(e) STROTSS]
  {\includegraphics[width=0.11\linewidth]{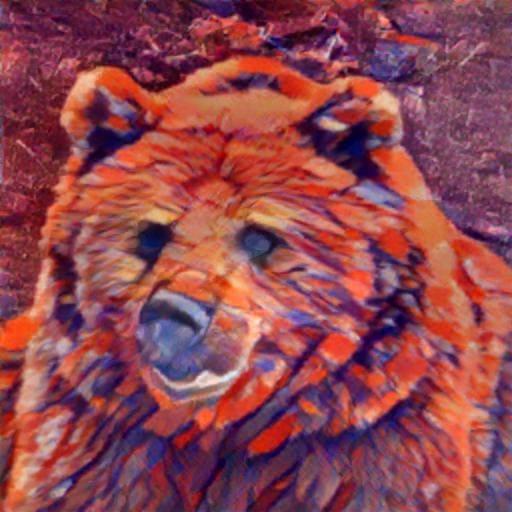}}\hfill
  \subfigure[(f) SinIR]
  {\includegraphics[width=0.11\linewidth]{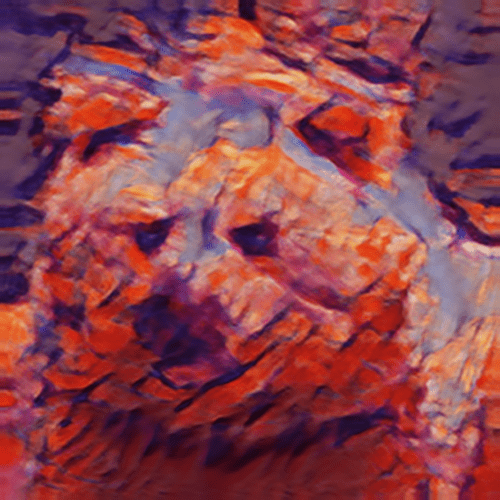}}\hfill
  \subfigure[(g) DTP]
  {\includegraphics[width=0.11\linewidth]{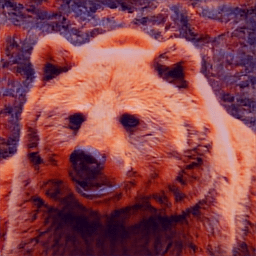}}\hfill
  \subfigure[(h) SANet]
  {\includegraphics[width=0.11\linewidth]{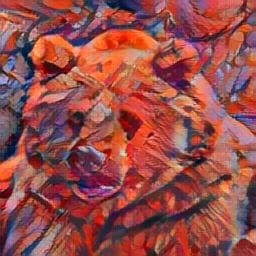}}\hfill
  \subfigure[(i) AdaAttn]
  {\includegraphics[width=0.11\linewidth]{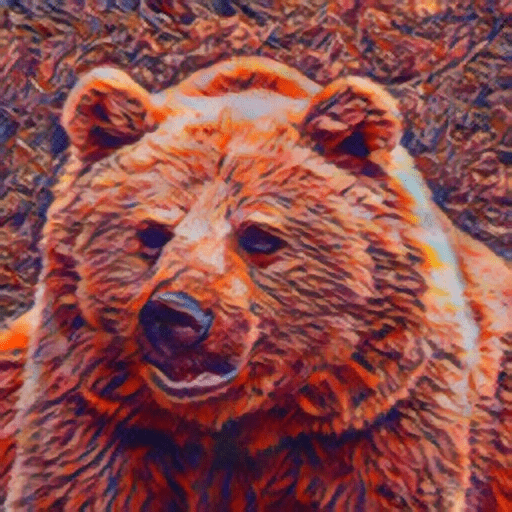}}\hfill\\
  

  \vspace{-10pt}
  \caption{\textbf{Qualitative comparison of existing artistic style transfer methods to our framework.}
  Our model shows more plausible style transfer results, especially in the respective of texture in the style image, in comparison with state-of-the-art methods.
  }

  \label{fig:qualitative}
\end{figure*}

\subsection{Controllable Style Transfer at Inference}
In this section, we introduce the methodology for the pixel-wise style manipulation and the resolution control that can be performed at inference time. With the test-time trained network, we can perform multiple application tasks such as region-wise style control, applying gradation effect, and resolution control without further training. 

In general, we can define different interpolated latent vectors to each coordinate point $p (x,y)$ as $z'_{p}$ by setting the interpolation rate $\alpha_{p}$ differently such that $z_{p}' = (\alpha_{p})z_{c} + (1-\alpha_{p})z_{s}$. The latent vectors and encoded vectors of coordinate positions at inference are denoted as $Z_{p}'$ and $E_{p}'$, respectively. Therefore, the inference stage can be defined as $I_{o} = \mathcal{D}(Z_{p}',E_{p}';\omega^{*})$. To control the resolution of $I_{o}$, we simply rearranged the dimension of input coordinates to $H_\mathrm{inf} \times W_\mathrm{inf}$ which are desirable height and width sizes.

\section{Results}
\subsection{Experimental Settings}
We empirically set the loss weights as $\lambda_\mathrm{cont} = 1$ and $\lambda_\mathrm{style} = 10^{5}$. The network is test-time trained on 256 $\times$ 256 resolution images by Adam optimizer ~\cite{kingma2014adam} with the learning rate of $10^{-3}$. All results of our model in experiments were inferred after test-time training without the additional learning process. We basically use the MS-COCO~\cite{lin2014microsoft} and WikiArti~\cite{phillips2011wiki} dataset for the content and style images respectively. All experiments were conducted using Intel i7-10700 CPU @ 2.90GHz, and NVIDIA RTX 3090 GPU.



\begin{table*}[t!]
    \centering

    \begin{tabular}{lcccccc}
    \toprule
            Resolution      & SANet             & AdaAttn           & URST          & Gatys et al.      & STROTSS   & Ours\\
    \midrule
        $256\times256$& 2.1/9.7h/0.056s & 19.0/5.7h/\textbf{0.014s} & 3.2/4.4h/1.55s   & 3.0/\textbf{6.48s}/-   & 2.8/2.8m/- & \textbf{1.7}/6.5m/0.25s \\
        $512\times512$ & 2.5/-/0.073s   & 20.5/-/\textbf{0.026s}     & 3.2/-/1.59s       & 3.8/\textbf{8.32s}/-   & 3.5/3.6m/-    & \textbf{1.7}/-/0.28s \\
        $2000\times2000$   &      $\times$        &     $\times$        & 4.0/-/2.76s      &  21.9/\textbf{1.1m}/-  & 22.5/8.4m/-   & \textbf{1.8}/-/0.96s \\
        $20000\times20000$ &      $\times$        &     $\times$        & 4.0/-/\textbf{63.4s}      &     $\times$        &     $\times$         & \textbf{1.8}/-/65.79s \\
    \bottomrule
    \end{tabular}
    \vspace{-5pt}
    \caption{\textbf{Memory usage and speed comparisons.} Each item means `memory usage (GiB) at inference time / learning time / inference time'. These comparisons are conducted on a single RTX 3090. We used AdaIN~\cite{huang2017arbitrary} as the baseline of URST~\cite{chen2022towards}. $\times$ means out of memory. }
    \label{tab:speed}
\end{table*}

        
\subsection{Qualitative Results}
In Fig.~\ref{fig:qualitative}, we compared our model with state-of-the-art methods. While our network yields a stable performance, existing methods show the results with insufficiently preserved content~\cite{kolkin2019style, park2019arbitrary, gatys2016image}, a lack of style ~\cite{kolkin2019style, yoo2021sinir, kim2021deep}, and undesirably deformed style~\cite{liu2021adaattn}. The learning-based methods, such as SANet~\cite{park2019arbitrary} and AdaAttn~\cite{liu2021adaattn}, failed to preserve content information with deformed style information for some arbitrary pair. The optimization-based methods showed better performance which were resulted from pair specific style transfer. However, the methods which optimize image itself~\cite{gatys2016image, kolkin2019style} displayed sub-optimal outcomes since they do not learn a prior for the specific image pair. To solve this problem, ~\cite{yoo2021sinir, kim2021deep} optimized the network via implicit and explicit approaches, respectively. Since these frameworks are designed for various generative applications and photorealistic style transfer, they can not make satisfactory results. In contrast, our framework is designed for artistic style transfer by optimizing the INR based network for specific image pair and shows better performance.

\subsection{Quantitative Results}
\begin{table}[t]
    \centering
    \vspace{-5pt}
    \scalebox{0.75}{
    \begin{tabular}{lcccc}
    \toprule
    Methods ($\alpha$)  & SANet ($0/1$) & AdaAttn ($0/1$) & Ours ($0/1$) & GT \\
    \midrule
            GRAM ($\downarrow/\uparrow$)   & 0.72/5.94   & 1.33/5.03  & \textbf{0.31}/\textbf{5.42} & 0.00/5.47\\ 
            SSIM ($\downarrow/\uparrow$)   & 0.39/0.64   & 0.56/0.62  & \textbf{0.34}/\textbf{0.91} & 0.13/1.00\\
    \bottomrule
    \end{tabular}}
    \vspace{-5pt}
    \caption{\textbf{Content preservation and stylization.} }
    \label{tab:quan}
\vspace{-5pt}
\end{table}

\paragraph{Memory Usage and Inference Speed.} In Table~\ref{tab:speed}, we compared the memory usage and inference speed of our framework with conventional algorithms~\cite{gatys2016image, kolkin2019style, park2019arbitrary,liu2021adaattn,chen2022towards}. Firstly, our network trained on an image with the size of 256 $\times$ 256. Then during inference, thanks to the properties of INR, our model shows significantly lower memory usage compared to other methods. After a short period of test-time training, our model can synthesize images of arbitrary resolutions at high inference speed, while the other methods can generate images of limited sizes.\vspace{-5pt}

\paragraph{Content Preservation and Stylization.} 
To provide a more detailed analysis of our model's performance, we estimated the stylization distance and content similarity with GRAM matrix distance and structural similarity index measure (SSIM), respectively. We compared our network with the other models that can interpolate between content and style at the inference stage to assess their capability for content preservation and stylization. For each sample out of 100 randomly sampled images, we obtained the scores for both cases where $\alpha$ are 0 or 1, and then averaged them. When $\alpha$ is 0, the model is expected to obtain low values in both metrics as it should be close to the style image in second-order statistics, while remaining far apart from a content image. When $\alpha$ is 1, the outcome should be similar to the content image which means both scores are expected to be close to thel GT score. In Table~\ref{tab:quan}., we can observe that AdaAttn~\cite{liu2021adaattn} and SANet~\cite{park2019arbitrary} achieve convincing scores, but our framework shows better results in both cases of $\alpha=0, 1$.

\begin{table}[t!]
    \centering
    \vspace{-5pt}
    \scalebox{0.9}{
    \begin{tabular}{lcccc}
    \toprule
    Method  & SANet & AdaAttn & Ours\\
    \midrule
            $d=1$   & 0.501   & 0.798  & \textbf{0.326}\\ 
            $d=3$   & 0.463   & 0.706  & \textbf{0.136}\\
    \bottomrule
    \end{tabular}}
    \vspace{-5pt}
    \caption{\textbf{Pixel-wise controllability.} }
    \label{tab:pixel}\vspace{-10pt}
\end{table}

\paragraph{Pixel-wise controllability.} 
We provide qualitative results on our model's ability to precisely apply the desired degree of style to each target pixel of an image without affecting neighboring pixels. We masked out a pixel $(i,j)$ of the image $I_o$ generated by the MLP generator and assigned style only to this particular pixel. Then, to verify if the style is applied only to this pixel and does not affect other pixels around it, we measured L1 
distance between neighboring pixels that are $d$ pixels away from the target pixel and the corresponding pixels in the same position on the reconstructed content image. It can be indicated that the smaller the distance value, the more accurately the model stylizes only the specific pixel of $I_o$ while less affecting the neighboring pixels. As our model showed a considerably lower value than other models, as shown in Tab.~\ref{tab:pixel}, we confirmed that compared to previous models, ours can more precisely assign the desired degree of style to each pixel of the image.


\subsection{Ablation Study}

\paragraph{Comparing Style Weights.} We observed  how the quality of the stylized output changes along the adjustment of style loss weight $\lambda_{\mathrm{style}}$. As shown in Fig.~\ref{styleweight}, the model produced the best outcome when $\lambda_{\mathrm{style}}=10^{5}$. In addition, we could also observe that the model achieves the best style controllability when $\lambda_{{\mathrm{style}}}=10^{5}$.
\vspace{-5pt}

\begin{figure}[t] \centering
    \renewcommand{\thesubfigure}{}
    
    \subfigure[]
    {\includegraphics[width=0.165\linewidth]{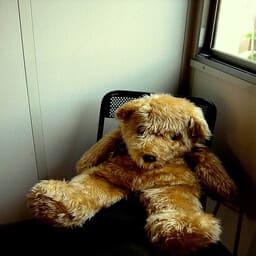}}\hfill    
    \subfigure[]
    {\includegraphics[width=0.165\linewidth]{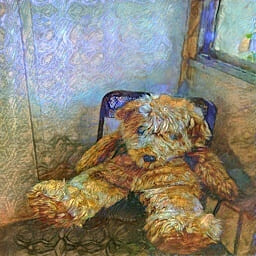}}\hfill
    \subfigure[]
    {\includegraphics[width=0.165\linewidth]{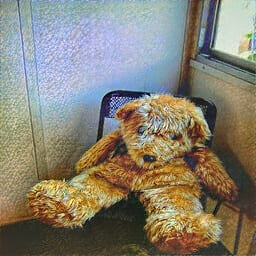}}\hfill
    \subfigure[]
    {\includegraphics[width=0.165\linewidth]{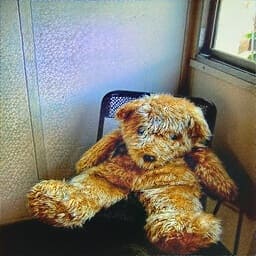}}\hfill 
    \subfigure[]
    {\includegraphics[width=0.165\linewidth]{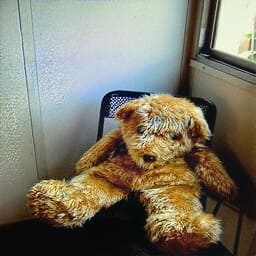}}\hfill 
    \subfigure[]
    {\includegraphics[width=0.165\linewidth]{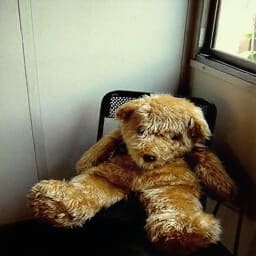}}\hfill \\
    \vspace{-22pt}

    \subfigure[]
    {\includegraphics[width=0.165\linewidth]{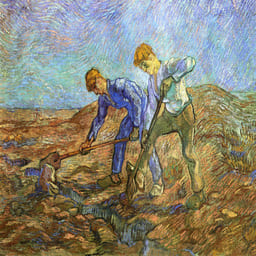}}\hfill 
    \subfigure[]
    {\includegraphics[width=0.165\linewidth]{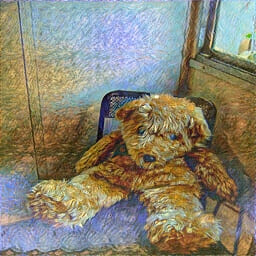}}\hfill 
    \subfigure[]
    {\includegraphics[width=0.165\linewidth]{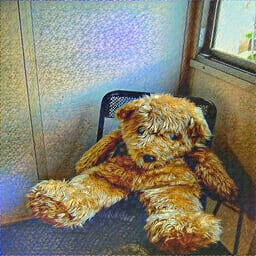}}\hfill 
    \subfigure[]
    {\includegraphics[width=0.165\linewidth]{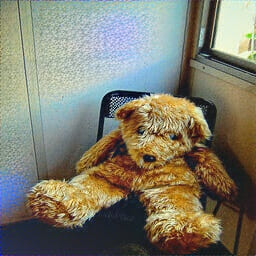}}\hfill 
    \subfigure[]
    {\includegraphics[width=0.165\linewidth]{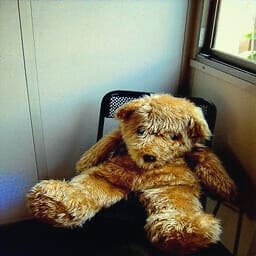}}\hfill 
    \subfigure[]
    {\includegraphics[width=0.165\linewidth]{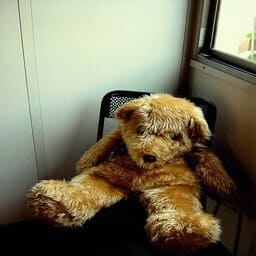}}\hfill \\
    \vspace{-20pt}
  \caption{\textbf{Loss function comparison.} We compared the effectiveness between loss function utilizing style loss function of ~\cite{gatys2016image} in our framework (top) and AdaIN~\cite{huang2017arbitrary} (bottom). The first column means content and style image. From the second to the last columns, images are inferred with $\alpha$ in a range of [0, 1].} 
\label{stylelosscompare}
\end{figure}

\paragraph{Comparing Style Losses.} According to SANet~\cite{park2019arbitrary}, the performance of the model may be different when different style losses are applied. We conducted an experiment to verify that there exists an effectiveness gap between these losses displayed in Fig.~\ref{stylelosscompare}. The experimental result does not indicate the significant difference in the model performance, and thus we argue that our framework works well with any style loss.\vspace{-10pt}

\begin{figure}[t] \centering
	\renewcommand{\thesubfigure}{}
    {\includegraphics[width=1\linewidth]{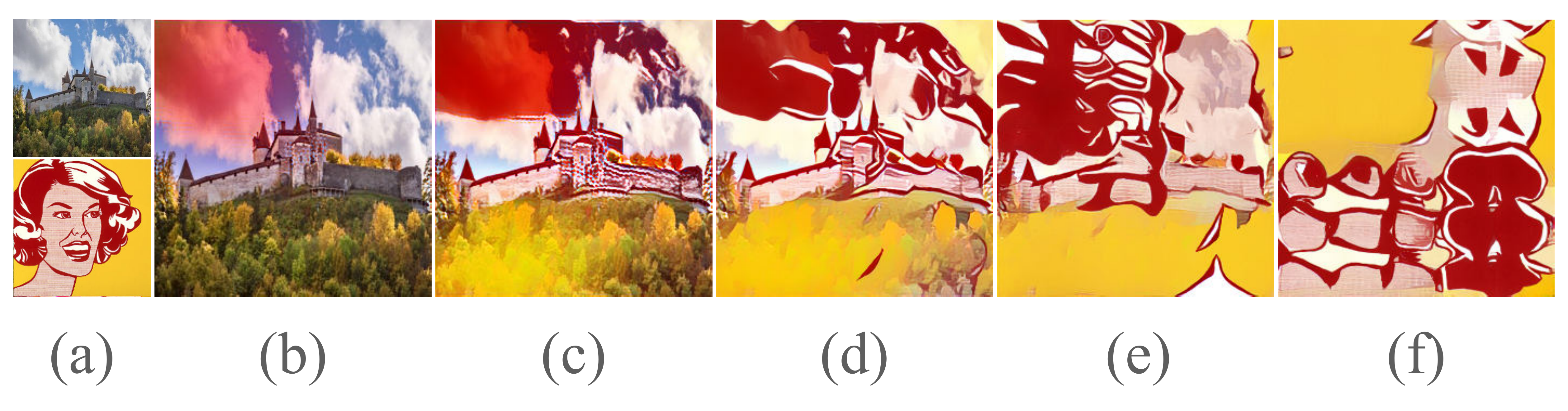}}\hfill\\
\vspace{-10pt}
  \caption{\textbf{Comparing style weights.} We tested the performance of the model by applying different values to the style weight $\lambda_{{\mathrm{style}}}$. The first column (a) means content and style image. The experimented results are shown from (b) to (f) of which the style weights $\lambda_{{\mathrm{style}}}$ are from $10^{3}$ to $10^{7}$. The model shows the best performance in terms of stylization quality and controllability when $\lambda_{{\mathrm{style}}}=10^{5}$. }
\label{styleweight}\vspace{-15pt}
\end{figure}

\begin{figure}[h] \centering
    \renewcommand{\thesubfigure}{}
    
	\subfigure[(a) w/ p.e.]    {\includegraphics[width=0.245\linewidth]{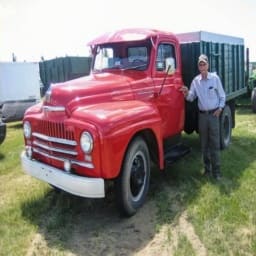}}\hfill
    \subfigure[(b) w/o p.e.]    {\includegraphics[width=0.245\linewidth]{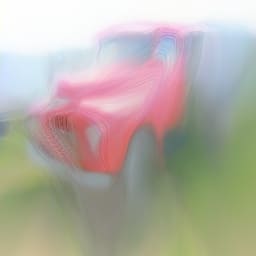}}\hfill
	\subfigure[(c) w/ p.e.]    {\includegraphics[width=0.245\linewidth]{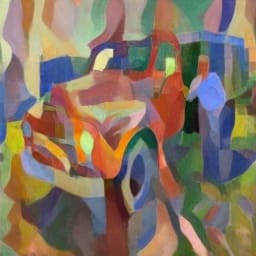}}\hfill
    \subfigure[(d) w/o p.e.]    {\includegraphics[width=0.245\linewidth]{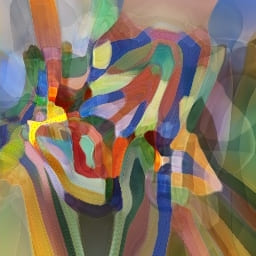}}\hfill
  \vspace{-10pt}
  \caption{\textbf{Positional encoding.} Positional encoding enables the model to generate high frequency details. (a), (b) are inferenced with $\alpha$ = 1 and (c), (d) are with $\alpha$ = 0.}
\label{positional}
\vspace{-10pt}
\end{figure}

\paragraph{Positional Encoding.} Recent study demonstrated that the positional encoding improves the network's ability to represent high frequency details of the image~\cite{mildenhall2020nerf}. Similarly to this, without positional encoding, the structure of the content image is significantly deformed at all ranges of $\alpha$ in our framework, as exemplified in Fig.~\ref{positional}.
\vspace{-3pt}

\section{Application}

\paragraph{Resolution Control.} 
Thanks to INR that represents features in continuous manner, we can arbitrarily manipulate an image resolution. Once our network is optimized on a pair of image with the resolution of $H \times W$, image of any scale can be generated by normalizing the dimension of the coordinates at inference $H_\mathrm{inf} \times W_\mathrm{inf}$ within the range of [-1,1]. For example, to generate $2,000 \times 3,000$ image, we divide the height and the width of continuous space that both have a range of [-1,1] by 2,000 and 3,000, respectively.\vspace{-8pt}

\begin{figure}[t] \centering
    \renewcommand{\thesubfigure}{}
    \subfigure[]
	{\includegraphics[width=1.\linewidth]{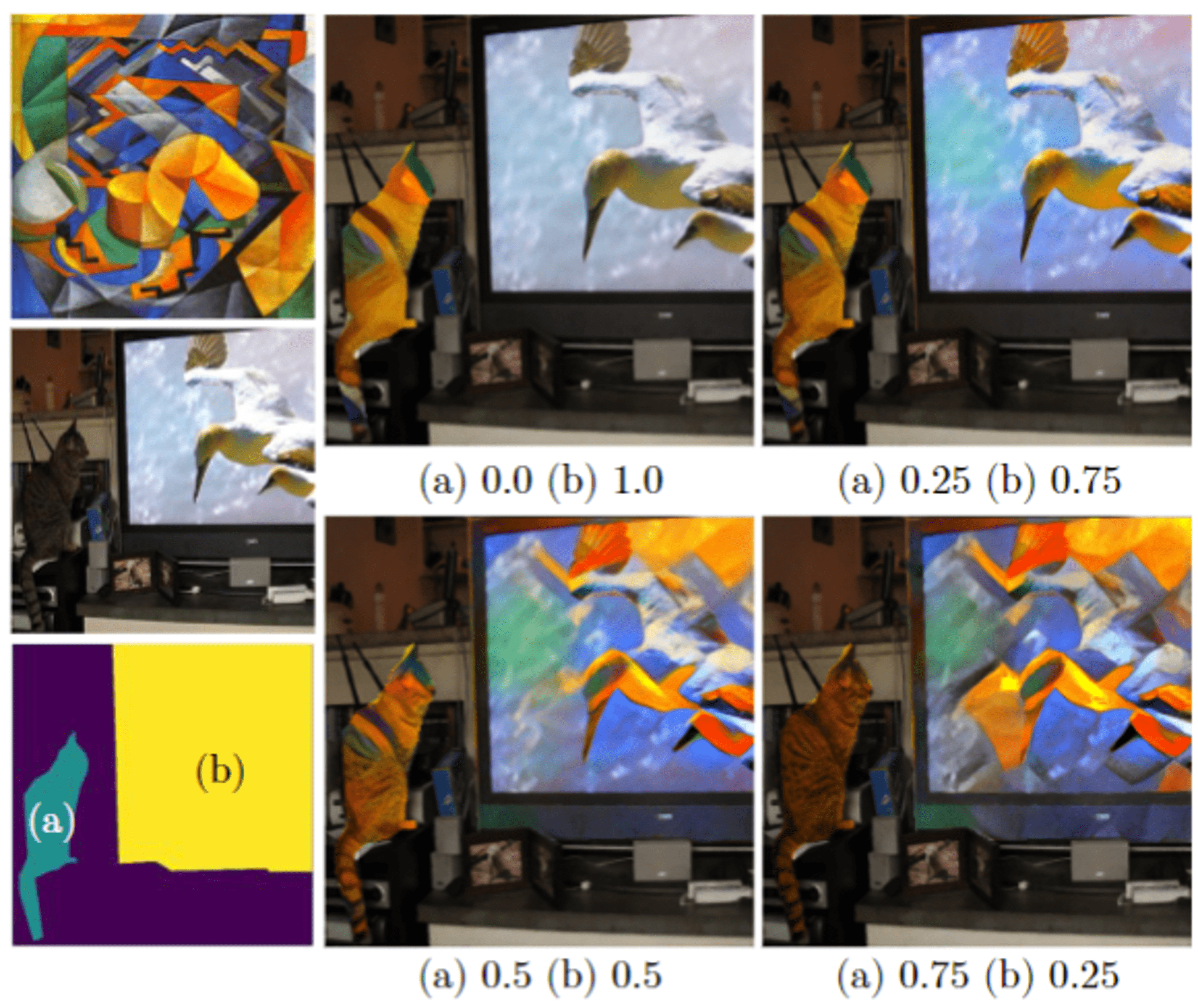}}\hfill
  \vspace{-15pt}
  \caption{\textbf{Spatial control using masks.} The first column shows style image, content image, and semantic mask respectively. The other columns show the result of stylization with different degrees to each mask.}
\label{fig:regionwise}
\vspace{-10pt}
\end{figure}

\paragraph{Pixel-wise control of stylization degree} Unlike existing CNN-based style transfer frameworks, our model can precisely apply the desired degree of style to each target pixel of an image, as displayed in Fig.~\ref{stylesplit}. We can observe that the regions are clearly distinguished at the boundary where the degree of style changes. This property is closely related to spatial control with mask in Fig~\ref{fig:regionwise}.
\vspace{-8pt}

 
\paragraph{Spatial control using masks.} We conducted an experiment to apply a different degree of style to different regions of the generated image with masks $M_{i} \in \bf{M}$. The different $\alpha$ is applied to obtain interpolated tensor $Z_{i}^{'}$ that corresponds to each mask. Next, the tensors are concatenated and provided to the network. We could modify distinct regions of the image with different degrees of style. (a) and (b) in Fig.~\ref{fig:teaser} and Fig~\ref{fig:regionwise}. show the region-wisely stylized outputs.

\begin{figure}[t] \centering
    \renewcommand{\thesubfigure}{}
    \subfigure[(a) Input]
	{\includegraphics[width=0.249\linewidth]{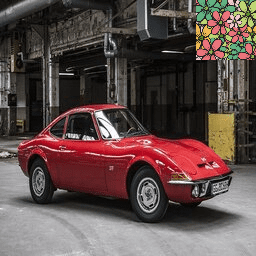}}\hfill
	\subfigure[(b) Ours]
    {\includegraphics[width=0.249\linewidth]{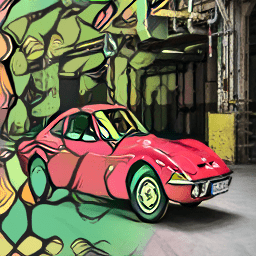}}\hfill
    \subfigure[(c) SANet]
    {\includegraphics[width=0.249\linewidth]{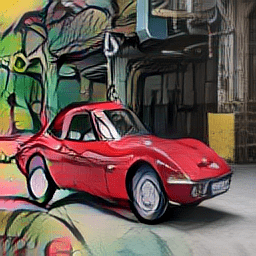}}\hfill
    \subfigure[(d) AdaAttn]
    {\includegraphics[width=0.249\linewidth]{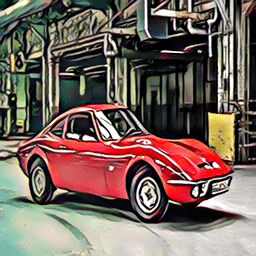}}\hfill\\
  
  \vspace{-10pt}
  \caption{\textbf{Gradation effect.} Our controllable style transfer network can apply gradation effect to the image. 
  }
  \label{gradation}
 \vspace{-10pt}
\end{figure}

\paragraph{Gradation Style Transfer Effect.} We can apply the gradation style transfer effect by continuously varying $\alpha$ in a single image. In Fig.~\ref{gradation} (b), from left to right, we changed $\alpha$ from 0 to 1 in column-wise manner. Although other existing models~\cite{park2019arbitrary,liu2021adaattn} can generate such effect, our method shows smoother and clear transition.

\begin{figure}[t]
    \centering
    \includegraphics[width=0.95\columnwidth]{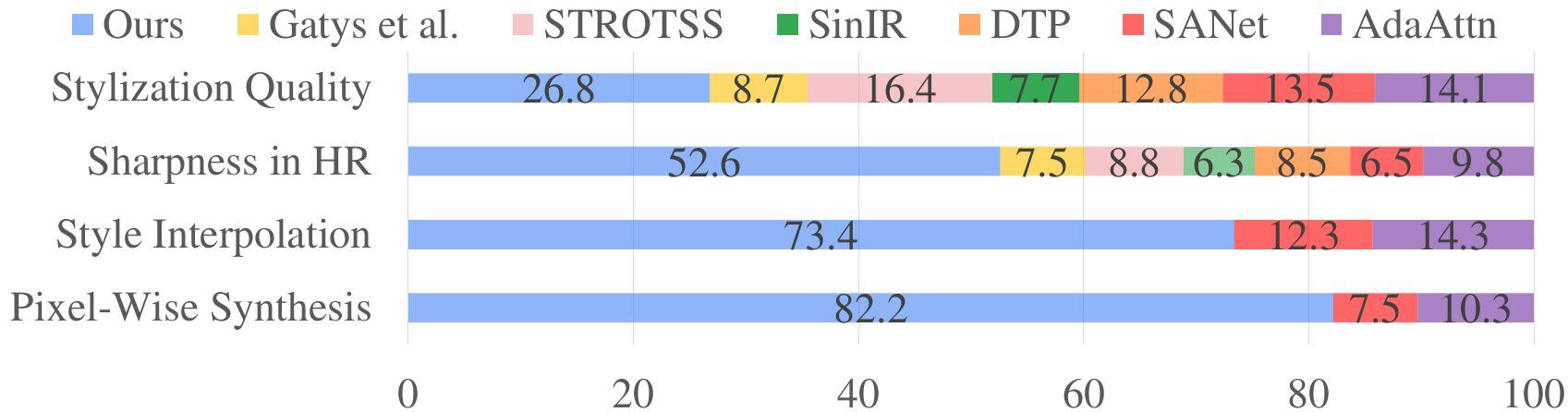}
    \vspace{-5pt}
    \caption{
    \textbf{User study results.} }
    \label{fig_userstudy}
\vspace{-10pt}
\end{figure}

\subsection{User Study.}
We conducted the user study on 1000 participants, a quantitative comparison that evaluates the performance of algorithms according to objective criteria. First, we asked each subject to indicate 30 images they prefer in aspect of stylization quality. Second, the participants were asked about the sharpness of the $2000\times2000$ resolution images generated by each model.  
Third, we asked the subjects to select the images that show smooth interpolation between style and content. Finally, to evaluate the models' ability to precisely transfer desired degree of style to specific part within an image, we asked participants to choose images that show clear separation at the boundary where degree of style changes as shown in Fig.~\ref{stylesplit}.\vspace{-5pt}

\section{Conclusion} 
In this paper, we propose, for the first time, a controllable style transfer framework that can pixel-wisely manipulate stylized images at test time. By learning image-pair specific prior and exploiting Implicit Neural Representation (INR), after test-time-trained once on an arbitrary image pair, our network can perform diverse applications of high-fidelity pixel-wise style transfer without additional training. Experimental results indicate that our model shows competitive performance in terms of stylization controllability, generation quality, and memory usage.

{\small
\bibliographystyle{ieee}
\bibliography{egbib}

\begin{thebibliography}{10}\itemsep=-1pt

\bibitem{atzmon2020sal}
Matan Atzmon and Yaron Lipman.
\newblock Sal: Sign agnostic learning of shapes from raw data.
\newblock In {\em CVPR}, 2020.

\bibitem{chabra2020deep}
Rohan Chabra, Jan~E Lenssen, Eddy Ilg, Tanner Schmidt, Julian Straub, Steven
  Lovegrove, and Richard Newcombe.
\newblock Deep local shapes: Learning local sdf priors for detailed 3d
  reconstruction.
\newblock In {\em ECCV}, 2020.

\bibitem{chen2021artistic}
Haibo Chen, Zhizhong Wang, Huiming Zhang, Zhiwen Zuo, Ailin Li, Wei Xing,
  Dongming Lu, et~al.
\newblock Artistic style transfer with internal-external learning and
  contrastive learning.
\newblock In {\em NeurIPS}, 2021.

\bibitem{chen2021dualast}
Haibo Chen, Lei Zhao, Zhizhong Wang, Huiming Zhang, Zhiwen Zuo, Ailin Li, Wei
  Xing, and Dongming Lu.
\newblock Dualast: Dual style-learning networks for artistic style transfer.
\newblock In {\em CVPR}, 2021.

\bibitem{chen2016fast}
Tian~Qi Chen and Mark Schmidt.
\newblock Fast patch-based style transfer of arbitrary style.
\newblock {\em arXiv preprint arXiv:1612.04337}, 2016.

\bibitem{chen2021learning}
Yinbo Chen, Sifei Liu, and Xiaolong Wang.
\newblock Learning continuous image representation with local implicit image
  function.
\newblock In {\em CVPR}, 2021.

\bibitem{chen2022towards}
Zhe Chen, Wenhai Wang, Enze Xie, Tong Lu, and Ping Luo.
\newblock Towards ultra-resolution neural style transfer via thumbnail instance
  normalization.
\newblock In {\em AAAI}, 2022.

\bibitem{chen2019learning}
Zhiqin Chen and Hao Zhang.
\newblock Learning implicit fields for generative shape modeling.
\newblock In {\em CVPR}, 2019.

\bibitem{dumoulin2016learned}
Vincent Dumoulin, Jonathon Shlens, and Manjunath Kudlur.
\newblock A learned representation for artistic style.
\newblock In {\em ICLR}, 2017.

\bibitem{gatys2015texture}
Leon Gatys, Alexander~S Ecker, and Matthias Bethge.
\newblock Texture synthesis using convolutional neural networks.
\newblock In {\em NeurIPS}, 2015.

\bibitem{gatys2016image}
Leon~A Gatys, Alexander~S Ecker, and Matthias Bethge.
\newblock Image style transfer using convolutional neural networks.
\newblock In {\em CVPR}, 2016.

\bibitem{gatys2017controlling}
Leon~A Gatys, Alexander~S Ecker, Matthias Bethge, Aaron Hertzmann, and Eli
  Shechtman.
\newblock Controlling perceptual factors in neural style transfer.
\newblock In {\em CVPR}, 2017.

\bibitem{gropp2020implicit}
Amos Gropp, Lior Yariv, Niv Haim, Matan Atzmon, and Yaron Lipman.
\newblock Implicit geometric regularization for learning shapes.
\newblock In {\em ICML}, 2020.

\bibitem{gu2018arbitrary}
Shuyang Gu, Congliang Chen, Jing Liao, and Lu Yuan.
\newblock Arbitrary style transfer with deep feature reshuffle.
\newblock In {\em CVPR}, 2018.

\bibitem{huang2017arbitrary}
Xun Huang and Serge Belongie.
\newblock Arbitrary style transfer in real-time with adaptive instance
  normalization.
\newblock In {\em ICCV}, 2017.

\bibitem{jiang2020local}
Chiyu Jiang, Avneesh Sud, Ameesh Makadia, Jingwei Huang, Matthias Nie{\ss}ner,
  Thomas Funkhouser, et~al.
\newblock Local implicit grid representations for 3d scenes.
\newblock In {\em CVPR}, 2020.

\bibitem{johnson2016perceptual}
Justin Johnson, Alexandre Alahi, and Li Fei-Fei.
\newblock Perceptual losses for real-time style transfer and super-resolution.
\newblock In {\em ECCV}, 2016.

\bibitem{kim2021deep}
Sunwoo Kim, Soohyun Kim, and Seungryong Kim.
\newblock Deep translation prior: Test-time training for photorealistic style
  transfer.
\newblock In {\em AAAI}, 2022.

\bibitem{kingma2014adam}
Diederik~P Kingma and Jimmy Ba.
\newblock Adam: A method for stochastic optimization.
\newblock In {\em ICLR}, 2015.

\bibitem{kolkin2019style}
Nicholas Kolkin, Jason Salavon, and Gregory Shakhnarovich.
\newblock Style transfer by relaxed optimal transport and self-similarity.
\newblock In {\em CVPR}, 2019.

\bibitem{li2016combining}
Chuan Li and Michael Wand.
\newblock Combining markov random fields and convolutional neural networks for
  image synthesis.
\newblock In {\em CVPR}, 2016.

\bibitem{li2017diversified}
Yijun Li, Chen Fang, Jimei Yang, Zhaowen Wang, Xin Lu, and Ming-Hsuan Yang.
\newblock Diversified texture synthesis with feed-forward networks.
\newblock In {\em CVPR}, 2017.

\bibitem{li2017universal}
Yijun Li, Chen Fang, Jimei Yang, Zhaowen Wang, Xin Lu, and Ming-Hsuan Yang.
\newblock Universal style transfer via feature transforms.
\newblock In {\em NeurIPS}, 2017.

\bibitem{li2018closed}
Yijun Li, Ming-Yu Liu, Xueting Li, Ming-Hsuan Yang, and Jan Kautz.
\newblock A closed-form solution to photorealistic image stylization.
\newblock In {\em ECCV}, 2018.

\bibitem{li2017demystifying}
Yanghao Li, Naiyan Wang, Jiaying Liu, and Xiaodi Hou.
\newblock Demystifying neural style transfer.
\newblock In {\em IJCAI}, 2017.

\bibitem{lin2014microsoft}
Tsung-Yi Lin, Michael Maire, Serge Belongie, James Hays, Pietro Perona, Deva
  Ramanan, Piotr Doll{\'a}r, and C~Lawrence Zitnick.
\newblock Microsoft coco: Common objects in context.
\newblock In {\em ECCV}, 2014.

\bibitem{liu2021adaattn}
Songhua Liu, Tianwei Lin, Dongliang He, Fu Li, Meiling Wang, Xin Li, Zhengxing
  Sun, Qian Li, and Errui Ding.
\newblock Adaattn: Revisit attention mechanism in arbitrary neural style
  transfer.
\newblock In {\em ICCV}, 2021.

\bibitem{lu2017decoder}
Ming Lu, Hao Zhao, Anbang Yao, Feng Xu, Yurong Chen, and Li Zhang.
\newblock Decoder network over lightweight reconstructed feature for fast
  semantic style transfer.
\newblock In {\em ICCV}, 2017.

\bibitem{mechrez2018contextual}
Roey Mechrez, Itamar Talmi, and Lihi Zelnik-Manor.
\newblock The contextual loss for image transformation with non-aligned data.
\newblock In {\em ECCV}, 2018.

\bibitem{mildenhall2020nerf}
Ben Mildenhall, Pratul~P Srinivasan, Matthew Tancik, Jonathan~T Barron, Ravi
  Ramamoorthi, and Ren Ng.
\newblock Nerf: Representing scenes as neural radiance fields for view
  synthesis.
\newblock In {\em ECCV}, 2020.

\bibitem{mordvintsev2018differentiable}
Alexander Mordvintsev, Nicola Pezzotti, Ludwig Schubert, and Chris Olah.
\newblock Differentiable image parameterizations.
\newblock {\em Distill}, 2018.

\bibitem{park2019arbitrary}
Dae~Young Park and Kwang~Hee Lee.
\newblock Arbitrary style transfer with style-attentional networks.
\newblock In {\em CVPR}, 2019.

\bibitem{park2019deepsdf}
Jeong~Joon Park, Peter Florence, Julian Straub, Richard Newcombe, and Steven
  Lovegrove.
\newblock Deepsdf: Learning continuous signed distance functions for shape
  representation.
\newblock In {\em CVPR}, 2019.

\bibitem{peng2020convolutional}
Songyou Peng, Michael Niemeyer, Lars Mescheder, Marc Pollefeys, and Andreas
  Geiger.
\newblock Convolutional occupancy networks.
\newblock In {\em ECCV}, 2020.

\bibitem{phillips2011wiki}
Fred Phillips and Brandy Mackintosh.
\newblock Wiki art gallery, inc.: A case for critical thinking.
\newblock {\em Issues in Accounting Education}, 26(3):593--608, 2011.

\bibitem{risser2017stable}
Eric Risser, Pierre Wilmot, and Connelly Barnes.
\newblock Stable and controllable neural texture synthesis and style transfer
  using histogram losses.
\newblock {\em arXiv preprint arXiv:1701.08893}, 2017.

\bibitem{Simonyan15}
Karen Simonyan and Andrew Zisserman.
\newblock Very deep convolutional networks for large-scale image recognition.
\newblock In {\em ICLR}, 2015.

\bibitem{sitzmann2020implicit}
Vincent Sitzmann, Julien Martel, Alexander Bergman, David Lindell, and Gordon
  Wetzstein.
\newblock Implicit neural representations with periodic activation functions.
\newblock In {\em NeurIPS}, 2020.

\bibitem{sitzmann2019scene}
Vincent Sitzmann, Michael Zollh{\"o}fer, and Gordon Wetzstein.
\newblock Scene representation networks: Continuous 3d-structure-aware neural
  scene representations.
\newblock In {\em NeurIPS}, 2019.

\bibitem{skorokhodov2021adversarial}
Ivan Skorokhodov, Savva Ignatyev, and Mohamed Elhoseiny.
\newblock Adversarial generation of continuous images.
\newblock In {\em CVPR}, 2021.

\bibitem{stanley2007compositional}
Kenneth~O Stanley.
\newblock Compositional pattern producing networks: A novel abstraction of
  development.
\newblock {\em Genetic programming and evolvable machines}, 2007.

\bibitem{tancik2020fourier}
Matthew Tancik, Pratul~P. Srinivasan, Ben Mildenhall, Sara Fridovich-Keil,
  Nithin Raghavan, Utkarsh Singhal, Ravi Ramamoorthi, Jonathan~T. Barron, and
  Ren Ng.
\newblock Fourier features let networks learn high frequency functions in low
  dimensional domains.
\newblock In {\em NeurIPS}, 2020.

\bibitem{ulyanov2016texture}
Dmitry Ulyanov, Vadim Lebedev, Andrea Vedaldi, and Victor~S Lempitsky.
\newblock Texture networks: Feed-forward synthesis of textures and stylized
  images.
\newblock In {\em ICML}, 2016.

\bibitem{ulyanov2016instance}
Dmitry Ulyanov, Andrea Vedaldi, and Victor Lempitsky.
\newblock Instance normalization: The missing ingredient for fast stylization.
\newblock {\em arXiv preprint arXiv:1607.08022}, 2016.

\bibitem{ulyanov2017improved}
Dmitry Ulyanov, Andrea Vedaldi, and Victor Lempitsky.
\newblock Improved texture networks: Maximizing quality and diversity in
  feed-forward stylization and texture synthesis.
\newblock In {\em ICCV}, 2017.

\bibitem{yoo2021sinir}
Jihyeong Yoo and Qifeng Chen.
\newblock Sinir: Efficient general image manipulation with single image
  reconstruction.
\newblock In {\em ICML}, 2021.

\bibitem{zhang2022cast}
Yuxin Zhang, Fan Tang, Weiming Dong, Haibin Huang, Chongyang Ma, Tong-Yee Lee,
  and Changsheng Xu.
\newblock Domain enhanced arbitrary image style transfer via contrastive
  learning.
\newblock In {\em ACM SIGGRAPH}, 2022.

\end{thebibliography}
}

\appendix
\onecolumn


\section{Network Structure Details}
\begin{figure}[h]
    \centering
    \includegraphics[width=1.0\columnwidth]{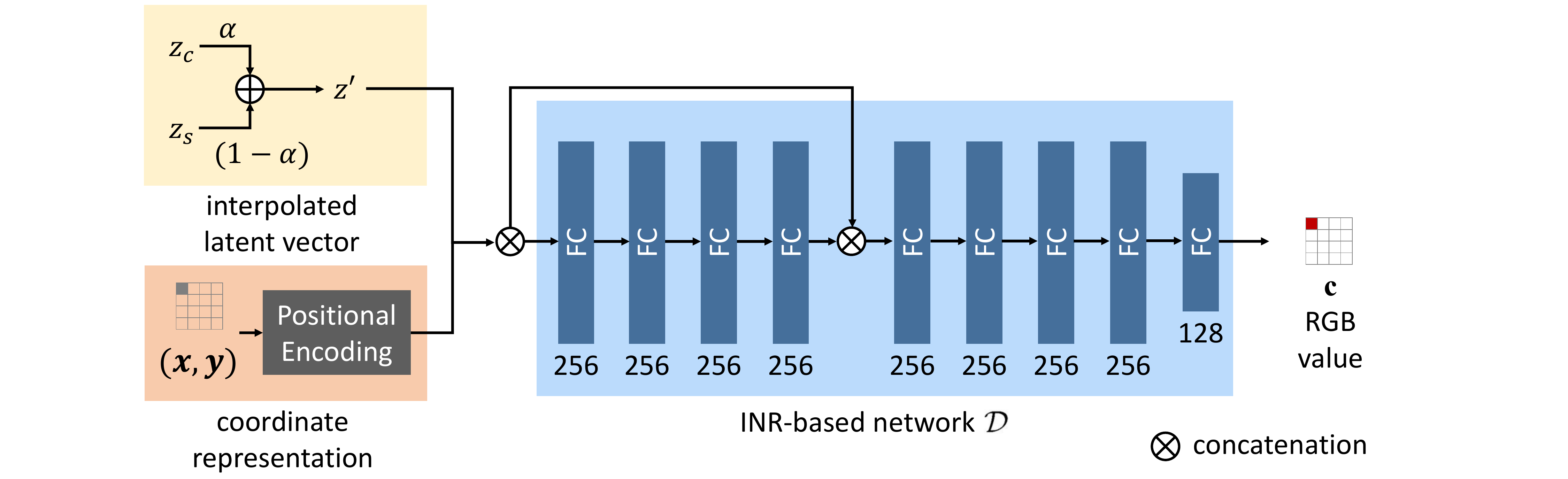}
    \vspace{-5mm}
    \caption{
    \textbf{Our INR-based network structure} The number below each fully connected layer means the number of features in that layer.
    }
    \label{networkdetail}
\end{figure}

We designed our network by referring to the structure of DeepSDF~\cite{park2019deepsdf}. It consists of 9 fully connected layers with ReLU activation attached to each layer. It takes two inputs; the interpolated latent vector $z' \in \mathbb{R}^{16}$ and the coordinate representation $\gamma(p)$ obtained from the positional encoding process. They are grouped into  $Z' \in \mathbb{R}^{HW \times 16}$ and $E$ respectively, and then given to the network as input after being concatenated. 
Meanwhile, the details of the positional encoding are as follows:
\begin{equation}
\gamma(p) = (sin(2^{0}\pi p), cos(2^{0}\pi p), \ldots , sin(2^{n_{f}-1}\pi p), cos(2^{n_{f}-1}\pi p))
\label{positionalencoding} \end{equation}
where $p \in \mathbb{R}^{2}$ denotes a single pixel coordinate which has a value within the range of [-1,1]. $n_{f}$ is set as 6 in our implementation. Additionally, we employed a pretrained VGG-19 network with fixed parameters as a feature extractor to test-time train our framework. 

\section{Additional Ablation Studies}

\begin{figure*}[t]
	\centering
	  {\includegraphics[width=1\linewidth]{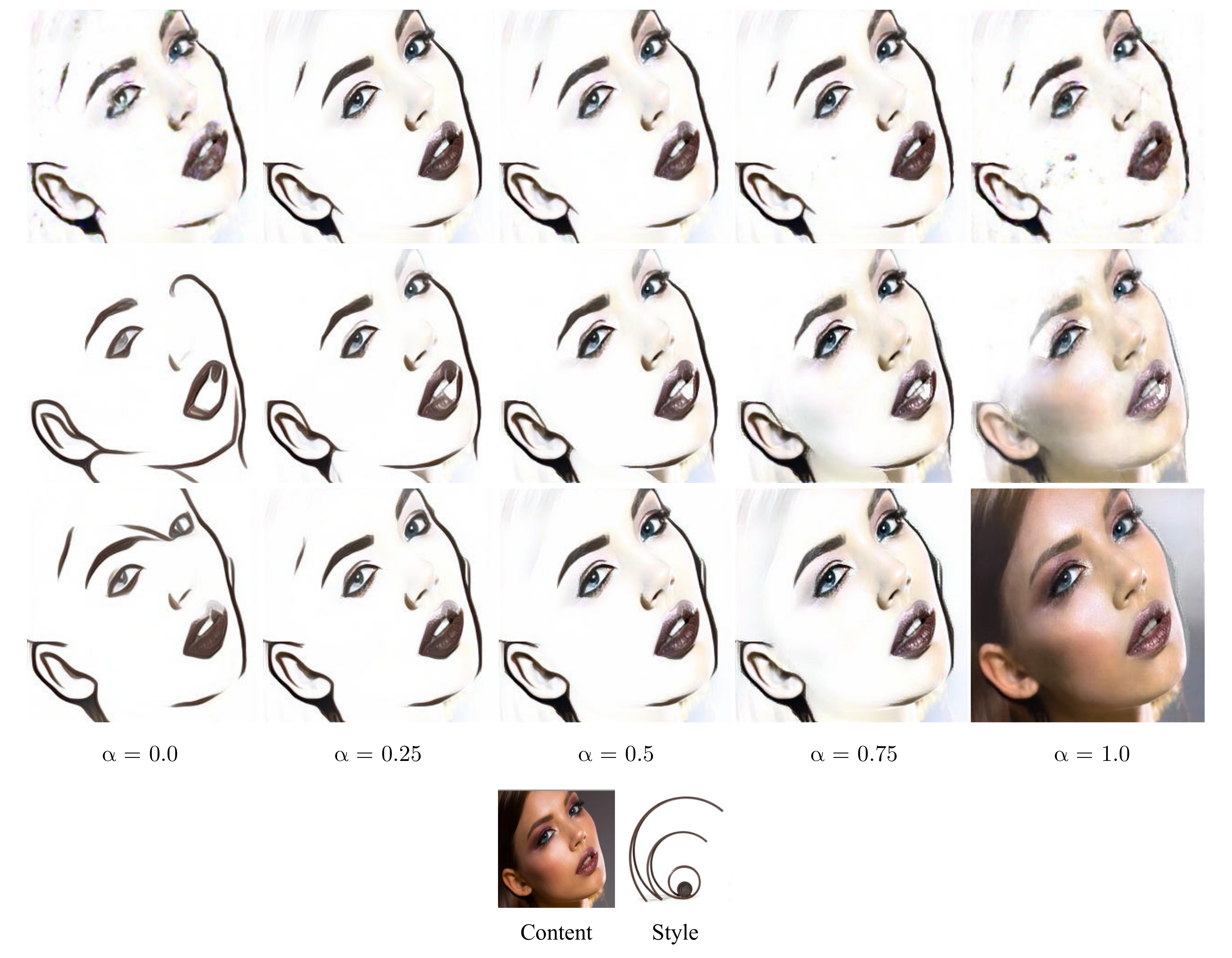}}\hfill

    \vspace{-10pt}

  \caption{\textbf{Sampling of $\alpha$} We compared the effectiveness of different sampling distributions. We could observe that the case of applying uniform distribution (third row) rather than fixed $\alpha$ (first row) or exponential distribution (second row) displays the best performance in terms of style controllability.  }
\label{alpha} \vspace{-9pt}
\end{figure*}

\paragraph{Sampling of $\alpha$.} As we described in the main text, section 4.2, using dynamically varying $\alpha$ as the interpolation ratio at each step is the key ingredient to achieving controllability in stylization. Therefore, here we experimented with some sampling distributions of $\alpha$ to find the optimal sampling method that maximizes the performance during test-time training in terms of content-style disentanglement and generation quality. In this experiment, uniform and exponential distributions of $\alpha$ and fixed $\alpha$ were tested. In the case of implementing exponential distributions, little modification is applied since $\alpha$ should be within the range of [0,1]. The experimental results are shown in Fig~\ref{alpha}. We could observe that sampling $\alpha$ from the uniform distribution yields the best results in terms of content-style disentanglement and generation quality.


\paragraph{Reweighting function.} Although we demonstrated the effectiveness of sampling $\alpha$ in a uniformly random manner in the previous section, as we mentioned in main text, section 4.2, exponentially reweighting $\alpha$ can further enhance the model's ability to make a distinct disentanglement between the style and the content. In order to prove that our exponential reweighting function improves the model's performance more than other reweighting functions, we compared several reweighting functions, including linear, polynomial, and exponential (ours) functions. Note that applying a linear function to $\alpha$ is the same as not reweighting the $\alpha$.

\begin{figure*}[t!] \centering
	\renewcommand{\thesubfigure}{}
	 {\includegraphics[width=1\linewidth]{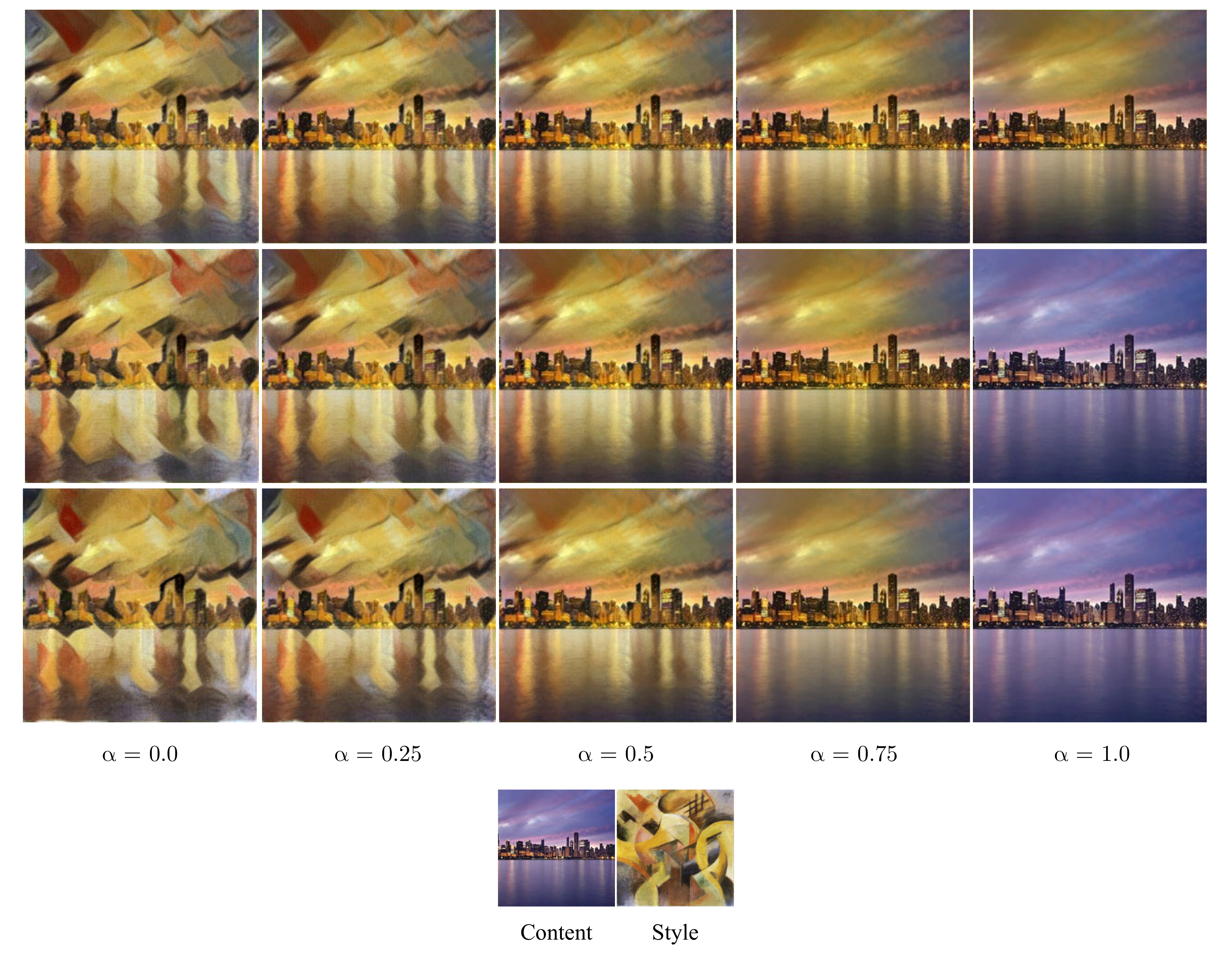}}\hfill
    \vspace{-10pt}
  \caption{\textbf{Effectiveness of exponential reweighting.} In this experiment, we observed that the model with an exponentially reweighting function (third row) rather than a linearly reweighting function (first row) or a polynomially reweighting function (second row) yields the best outcomes in terms of content-style disentanglement. }
\label{supplereweighting}
\end{figure*}

As shown in Fig~\ref{supplereweighting}, we observed that the model with an exponential reweighting function shows the best outcomes. We also compared the content-style interpolation results between the default model and the model with exponentially reweighted $\alpha$, and we could observe that the model with exponentially reweighted $\alpha$ achieves better disentanglement, as shown in Table.~\ref{tab:disentangle}.  

\paragraph{Comparing Style Weights.}We also conducted an experiment to find the optimal style weights $\lambda_\mathrm{style}$. The model shows the best performance in terms of stylization quality and controllability when $\lambda_\mathrm{style}=10^{5}$. Fig~\ref{styleweight} describes the details.

\begin{figure*}[h!]
    \includegraphics[width=1\textwidth]{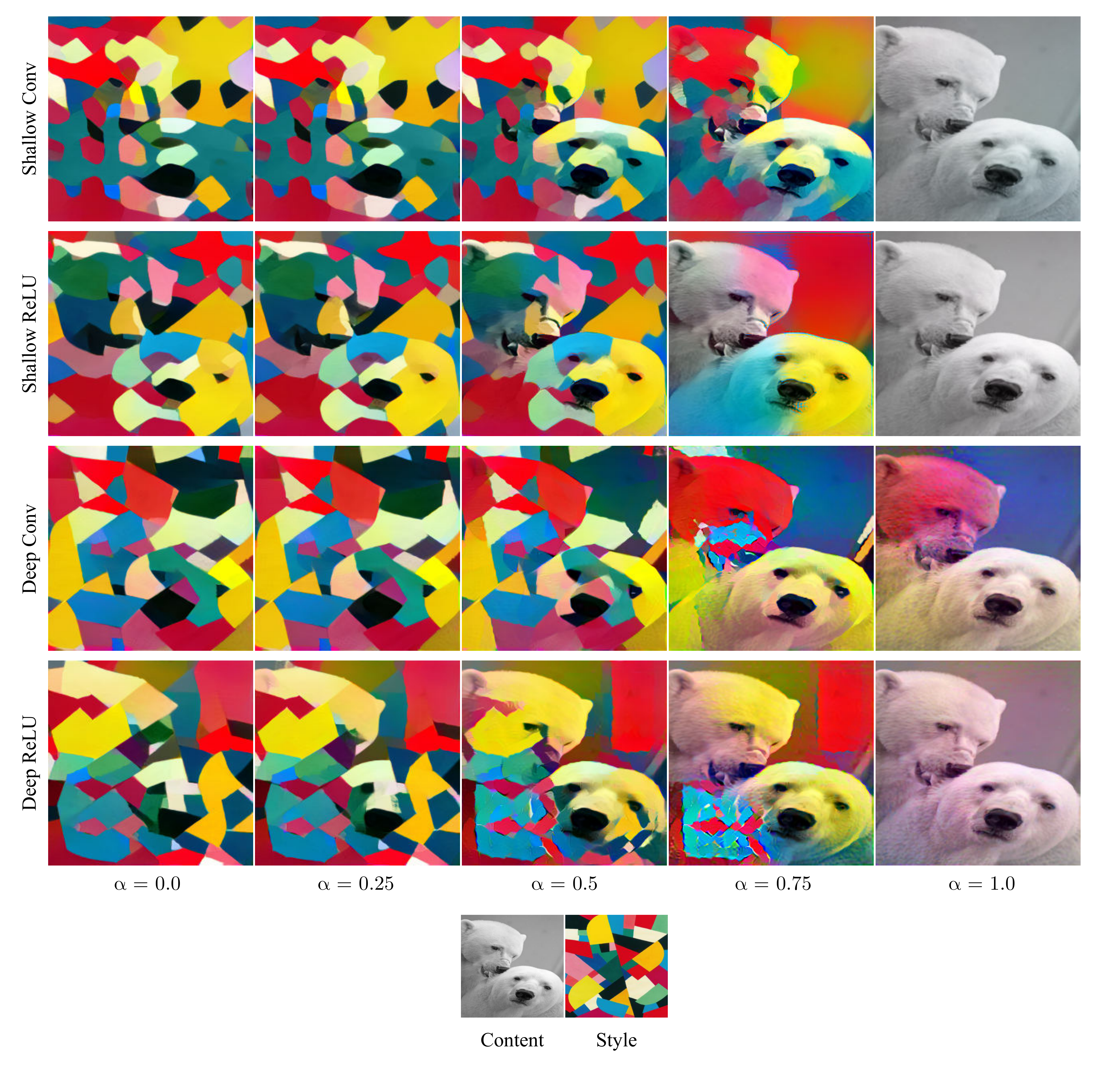}\hfill\\
 \vspace{-25pt}
  \caption{\textbf{Comparing the sets of VGG layers} Using feature maps from \emph{Shallow ReLU} (\textit{relu}1\_1, \textit{relu} 1\_3, \textit{relu}2\_2, \textit{relu}2\_4, \textit{relu}3\_2) shows the best results in terms of content preservation and quality of style transfer. In comparison with other cases, we can observe that the style intensity changes more smoothly with the increase in $\alpha$ value when these layers are used. Meanwhile, using the \emph{Deep Conv} or \emph{Deep ReLU} leads to failure in restoring the content image when $\alpha=1.0$}
\label{stylelayer}
\end{figure*}

\paragraph{Layer Selection for Style Loss Computation.} It is very important to decide which VGG feature map to use for style loss computation. There are two things to consider: which layers to use and whether to use feature maps that have passed through the activation function or feature maps that have not. We can compute the style loss on shallow VGG layers (conv1\_1, conv1\_3, conv 2\_2, conv2\_4, conv3\_2 of the VGG-19 model) or deep layers (conv1\_2, conv2\_3, conv3\_3, conv4\_3, conv5\_3). In this paper, these two cases are denoted as \emph{Shallow Conv} and \emph{Deep Conv}, respectively. We can also compute the loss on shallow VGG layers (\textit{relu}1\_1, \textit{relu}1\_3, \textit{relu}2\_2, \textit{relu}2\_4, \textit{relu}3\_2) or deep VGG layers (\textit{relu}\_2, \textit{relu}2\_3, \textit{relu}3\_3, \textit{relu}4\_3, \textit{relu}5\_3) after the activation (ReLU). In this paper, these two cases are denoted as \emph{Shallow ReLU} and \emph{Deep ReLU}, respectively. Therefore, there are four cases in total. Fig~\ref{stylelayer} shows the results of comparing these cases. We could observe that employing shallow layers after the activation (\textit{relu}1\_1, \textit{relu}1\_3, \textit{relu}2\_2, \textit{relu}2\_4, \textit{relu}3\_2) is the best in terms of content preservation and style intensity control.

\begin{table*}[h!]
    \centering

    \begin{tabular}{lccccccc}
    \toprule
         Method \textbackslash  $\alpha$  &   $\alpha=0.0$    & $\Delta\alpha_{(0.0,0.25)}$ & $\Delta\alpha_{(0.25,0.5)}$ & $\Delta\alpha_{(0.5,0.75)}$ & $\Delta\alpha_{(0.75,1.0)}$ & $\alpha=1.0$    \\
    \midrule
        Without reweighting                   & 27.98/0.02  & 0.01/0.04   &   0.01/0.08   &  0.06/0.21   &   0.44/1.56   &  28.49/1.76  \\
        With reweighting       & \bf{27.97/ 0.02}  & \bf{0.09/0.04}  & \bf{0.54/0.76}    &  \bf{3.46/2.62}  &  \bf{3.22/2.40}  & \bf{35.28/5.04}   \\
    \bottomrule
    \end{tabular}
    
    \caption{\textbf{Disentanglement comparisons.} {Each item means PSNR and a distance between gram matrices which measure the degree of content preservation and stylization respectively, according to different $\alpha$. The higher PSNR value means better preservation of content, and a close distance between gram matrices indicates better stylization. To measure disentanglement, we calculated the difference in metrics between two images with different $\alpha$, which is defined as $\Delta\alpha_{(a,b)}$. The larger gaps than with the default method show that reweighting $\alpha$ makes the network achieve better disentanglement.}}
    \label{tab:disentangle}
\end{table*}

\begin{figure*}[t!]
\includegraphics[width=1\textwidth]{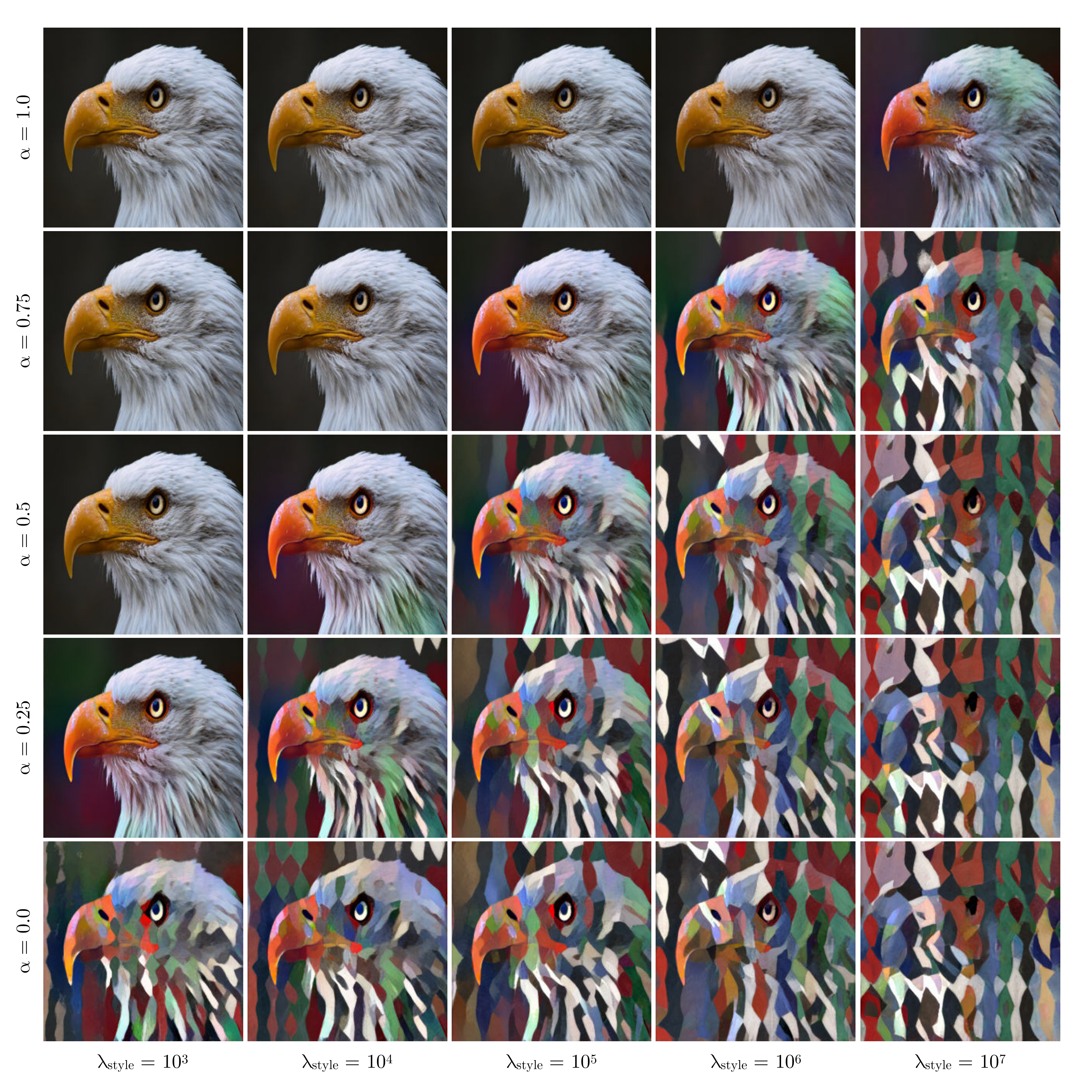}\hfill\\
 \vspace{-22pt}
  \caption{\textbf{Comparing style weights.} Not only the performance of the style transfer but also the style intensity change according to the adjustment of $\alpha$ was the most noticeable when $\lambda_{style}=10^{5}$. If the $\lambda_{style}$ value is too small, the change in style according to the adjustment of the $\alpha$ value is not revealed well and only the content tends to be well preserved. When $\lambda_{style}$ is too big, style changes are clearly observed, but the content is not preserved well. }
\label{styleweight}
\end{figure*}

\newpage

\begin{figure*}[t]
\centering
    \includegraphics[width=1\textwidth]{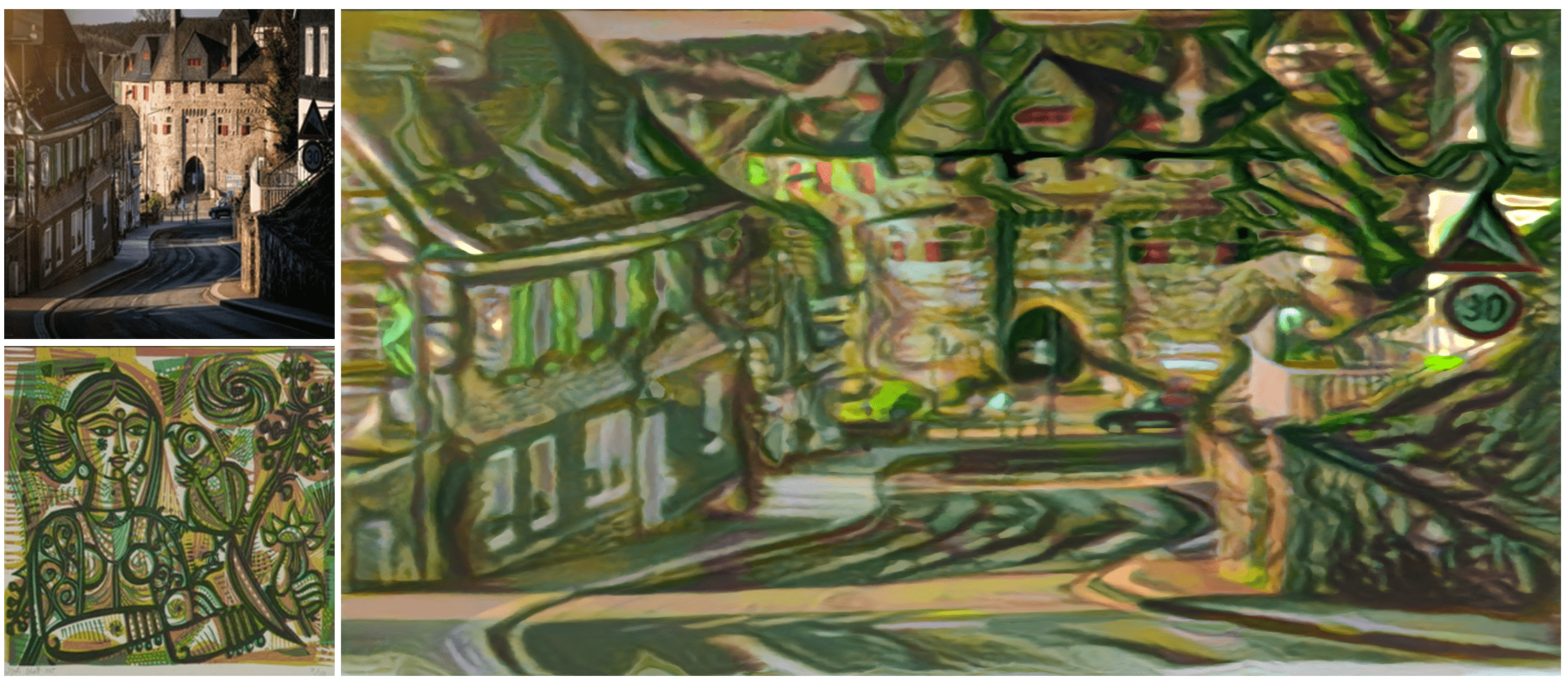}\hfill\\
    \vspace{-22pt}
    \includegraphics[width=1\textwidth]{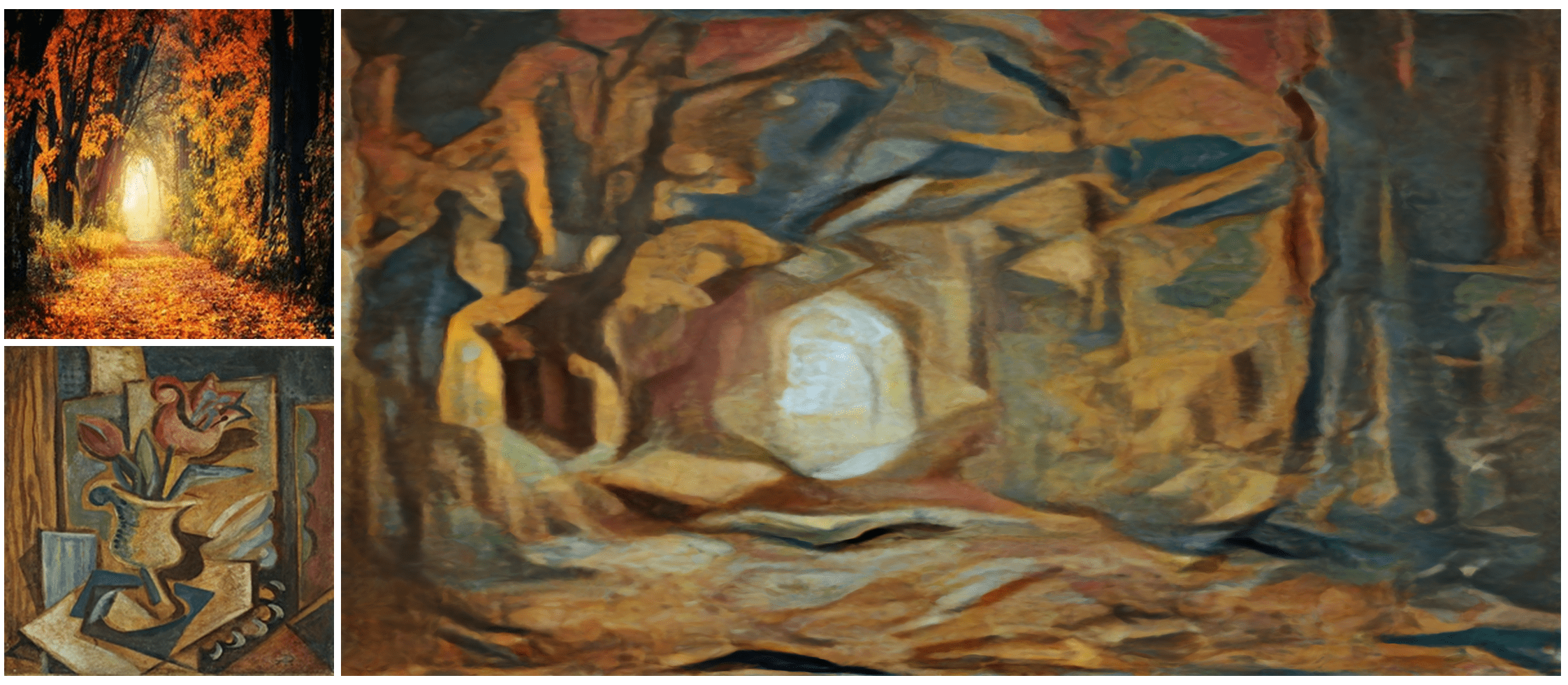}\hfill\\
    \vspace{-15pt}
    \includegraphics[width=1\textwidth]{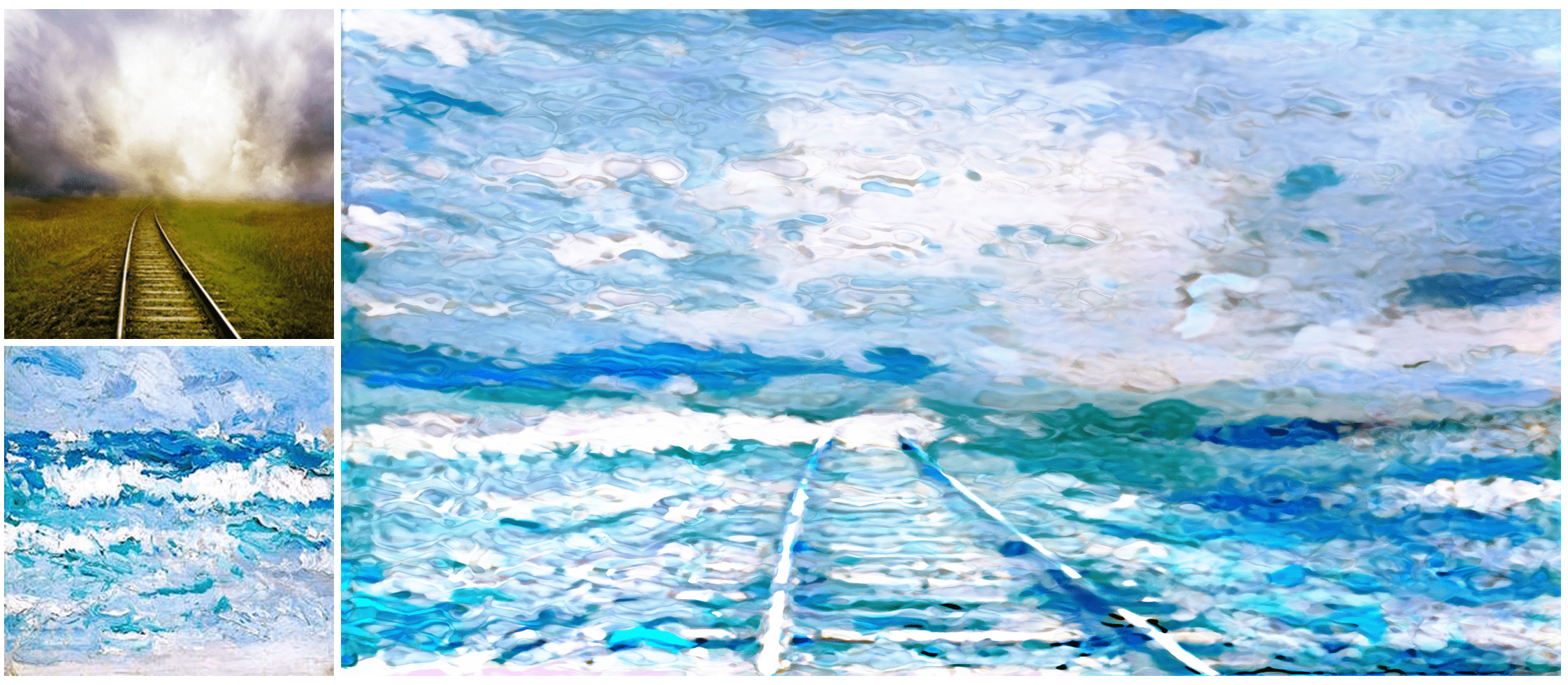}\hfill\\
    \vspace{-5pt}
\caption{\textbf{Resolution control.} Images with any resolution size and aspect ratio can be produced by our model. All of the images are generated by the network optimized on a 256 $\times$ 256 scale. The Above images have a size of 6000 $\times$ 12000 and are generated at inference time.}
\label{fig:resolution}
\end{figure*}

\clearpage
\newpage

\section{Additional Exemplars of Controllable Style Transfer}

\begin{figure*}[t]
\includegraphics[width=1\textwidth]{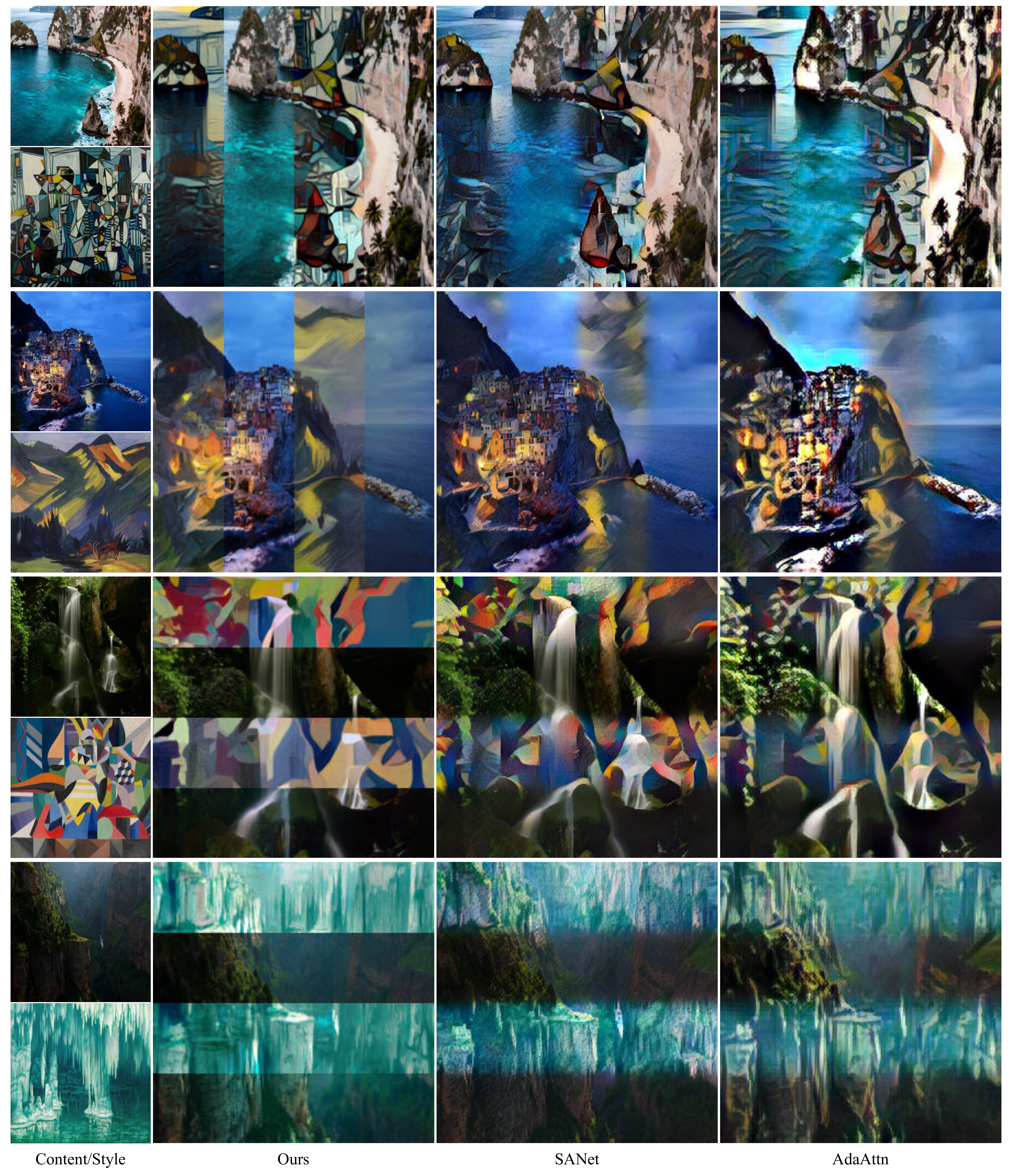}\hfill\\
\vspace{-15pt}

\caption{\textbf{Pixel-wise Synthesis.} In the image created by our model, we can observe that the regions are clearly distinguished at the point where the style intensity significantly changes, as our model can control the style of the image in a pixel-wise manner. However, CNN based methods~\cite{liu2021adaattn,park2019arbitrary} tend to show unclear separation at the boundary where style intensity changes. We applied different $\alpha$ values to the four different regions of each image created by three different models, including ours.}
\label{fig:split}
\end{figure*}

\paragraph{Pixel-wise synthesis.}
Our INR-based network is able to perform pixel-wise style transfer. Unlike existing CNN-based networks, our model can apply a different degree of style to each pixel of an image.  Fig~\ref{fig:split} shows examples of it.

\begin{figure*}[!t]
 \centering
 \includegraphics[width=1\textwidth]{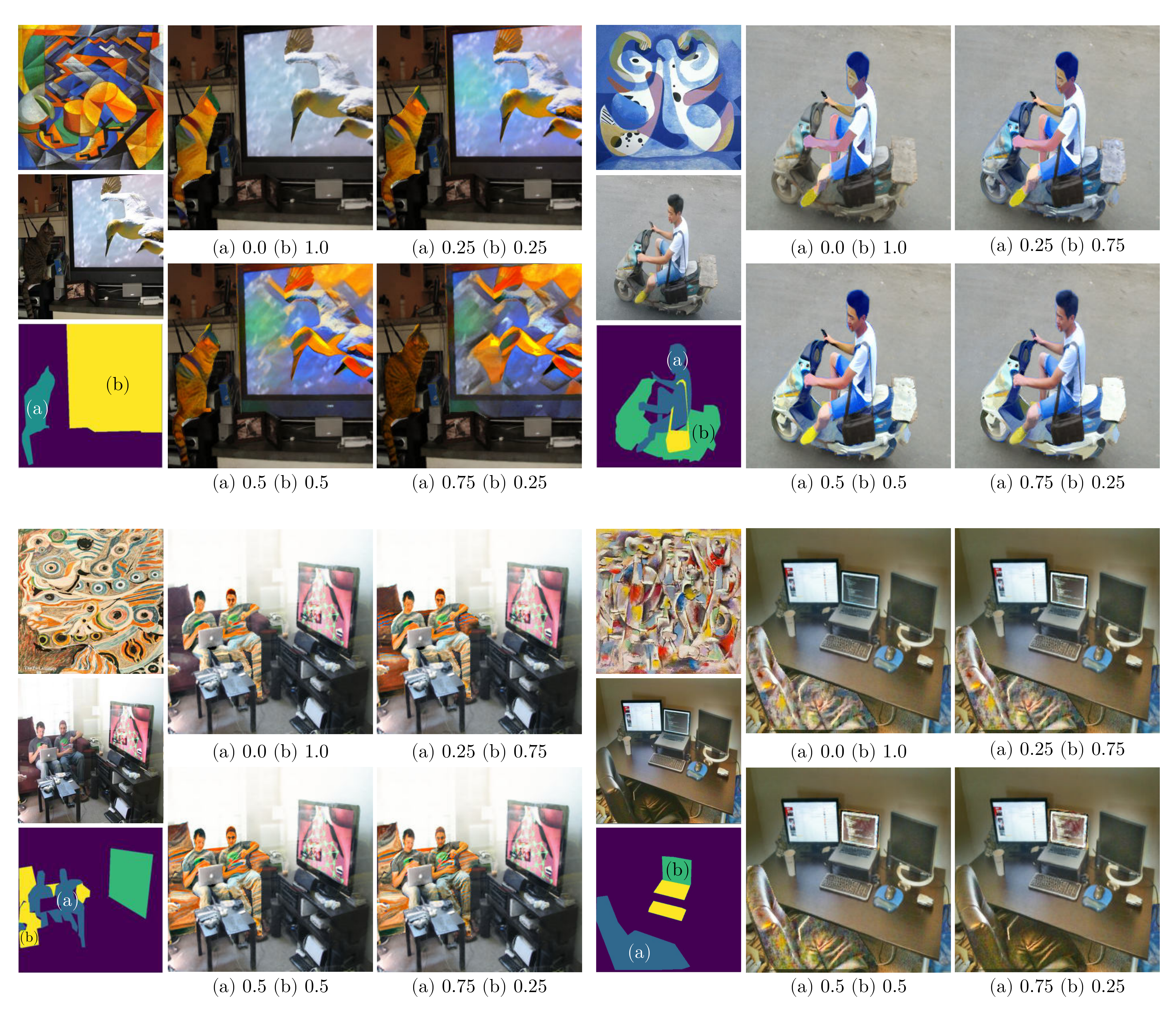}\hfill\\
  \vspace{-5pt}
  \caption{\textbf{Region-wise style transfer.} Our model can adjust the degree of style transfer by changing $\alpha$ at inference time after test-time training. Also, it is possible to apply different $\alpha$ to different regions. Each subcaption means the $\alpha$ value applied to each masked region.}
 \label{mask}
 \end{figure*}

\paragraph{Region-wise style transfer.}
As our network is able to pixel-wisely control the degree of stylization, it can also stylize each region of an image in an elaborate manner. Our region-wise style transfer results are shown in Fig~\ref{mask}. Furthermore, in Fig~\ref{fig:split}, we can observe that, unlike the image generated by the existing CNN-based model, the regions in the image generated by our model are accurately separated at the points where the style intensity is greatly changed.

\begin{figure*}[t]
\includegraphics[width=1\textwidth]{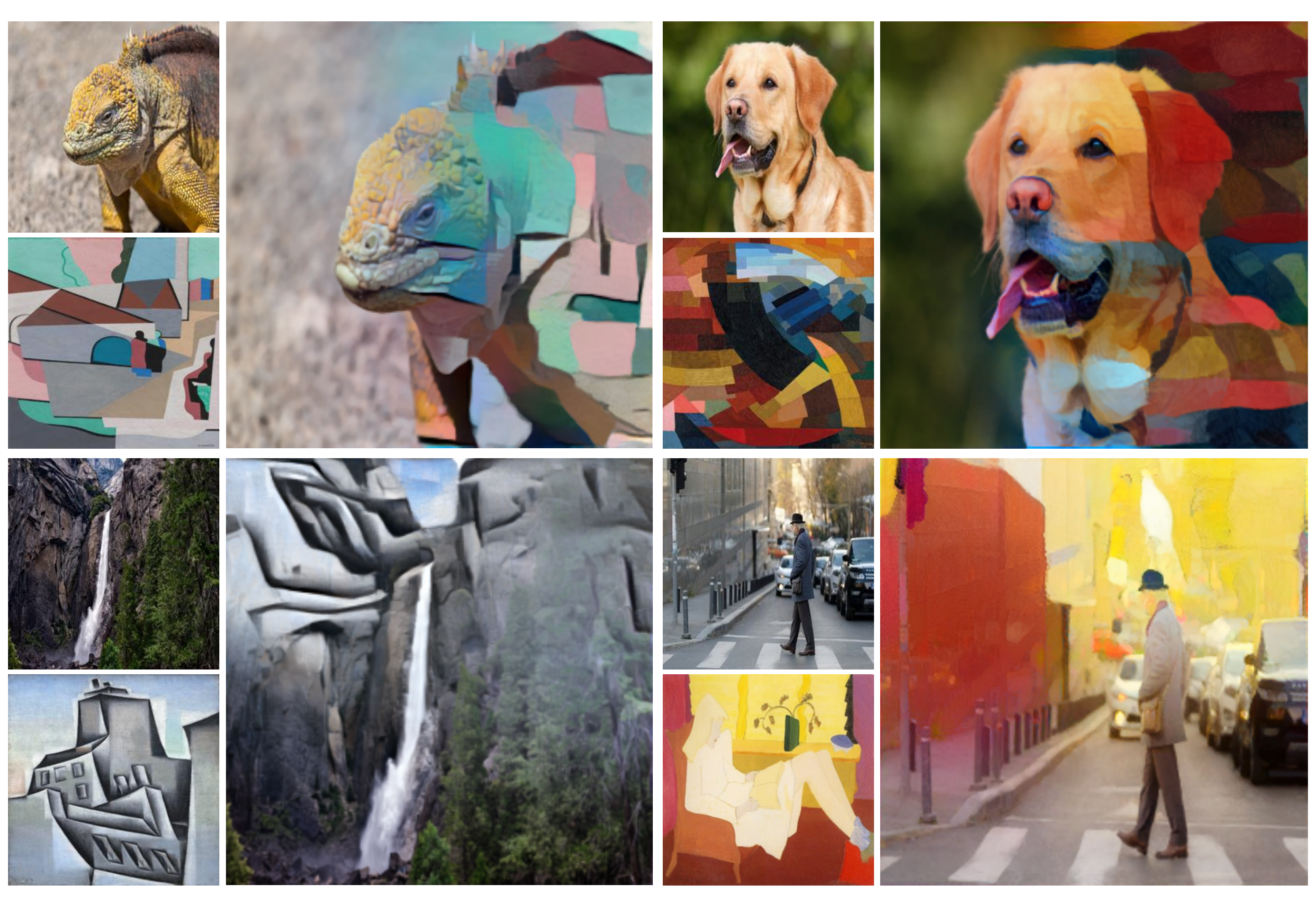}\hfill\\

\vspace{-15pt}
\caption
{\textbf{Gradation Effect.} Our controllable style transfer network is able to apply a gradation effect to the image.}
\label{fig:gradation}
\end{figure*}

\paragraph{Gradation effect.} Our model can give a gradation effect to the image by applying different degrees of style to each column or row of it. Examples are displayed in Fig~\ref{fig:gradation}. 

\begin{figure*}\centering
  \renewcommand{\thesubfigure}{}
  
  \subfigure[]  {\includegraphics[width=0.11\linewidth]{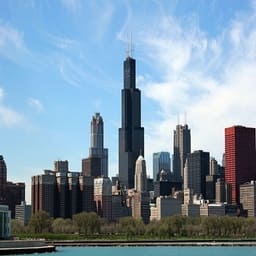}}\hfill
  \subfigure[]  {\includegraphics[width=0.11\linewidth]{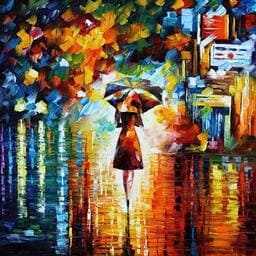}}\hfill
  \subfigure[]  {\includegraphics[width=0.11\linewidth]{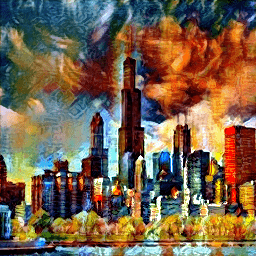}}\hfill
  \subfigure[]  {\includegraphics[width=0.11\linewidth]{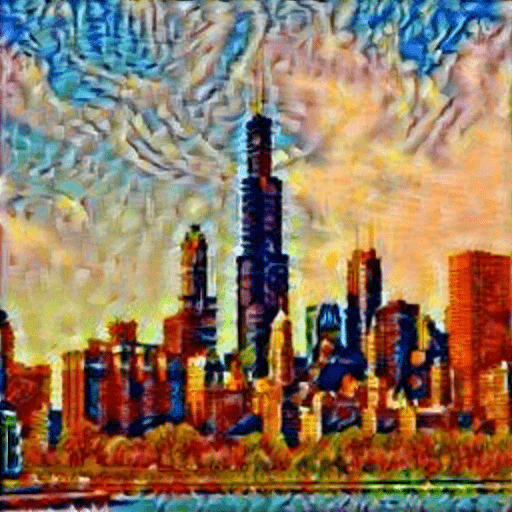}}\hfill
  \subfigure[]  {\includegraphics[width=0.11\linewidth]{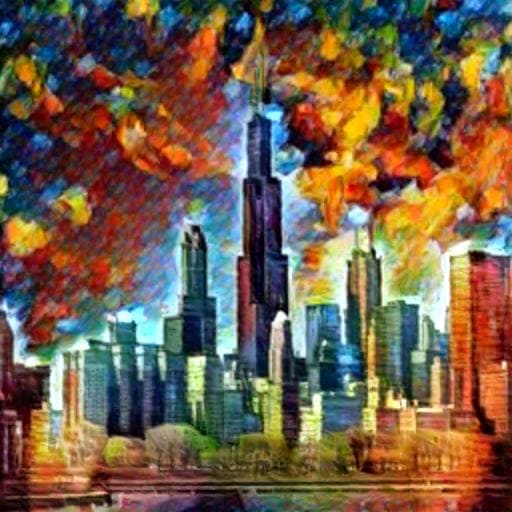}}\hfill
  \subfigure[]  {\includegraphics[width=0.11\linewidth]{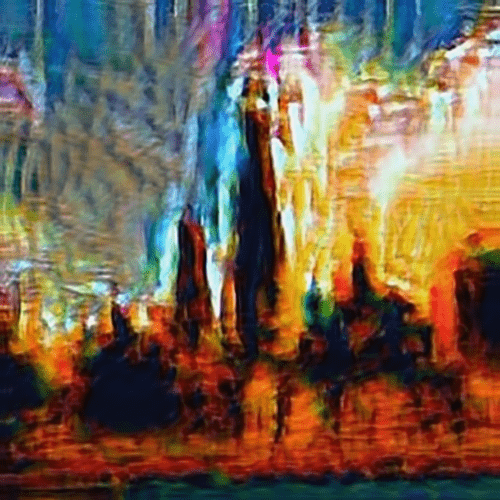}}\hfill
  \subfigure[]  {\includegraphics[width=0.11\linewidth]{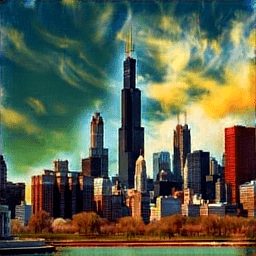}}\hfill
  \subfigure[]  {\includegraphics[width=0.11\linewidth]{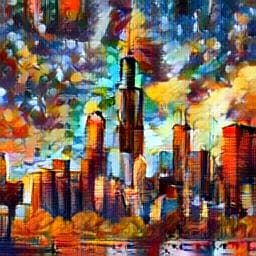}}\hfill
  \subfigure[]  {\includegraphics[width=0.11\linewidth]{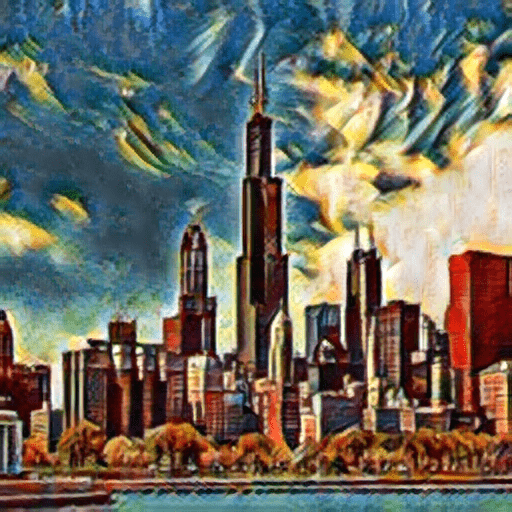}}\hfill\\
  \vspace{-22pt}
  
  \subfigure[]  {\includegraphics[width=0.11\linewidth]{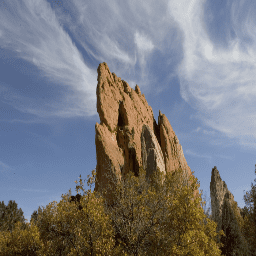}}\hfill
  \subfigure[]  {\includegraphics[width=0.11\linewidth]{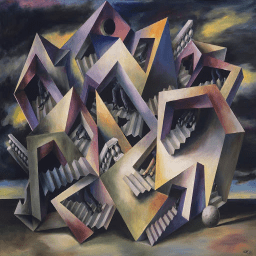}}\hfill
  \subfigure[]  {\includegraphics[width=0.11\linewidth]{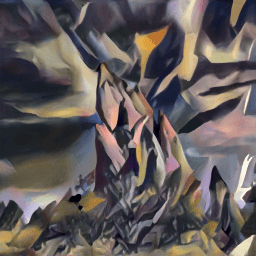}}\hfill
  \subfigure[]  {\includegraphics[width=0.11\linewidth]{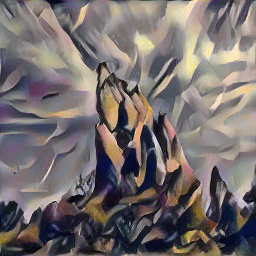}}\hfill
  \subfigure[]  {\includegraphics[width=0.11\linewidth]{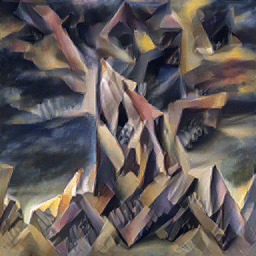}}\hfill
  \subfigure[]  {\includegraphics[width=0.11\linewidth]{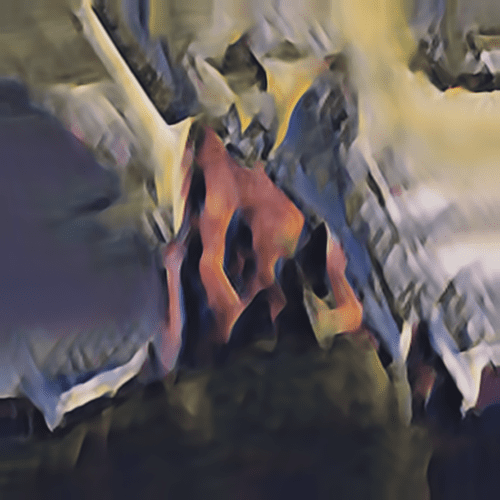}}\hfill
  \subfigure[]  {\includegraphics[width=0.11\linewidth]{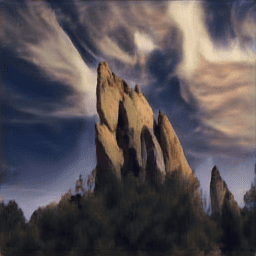}}\hfill
  \subfigure[]  {\includegraphics[width=0.11\linewidth]{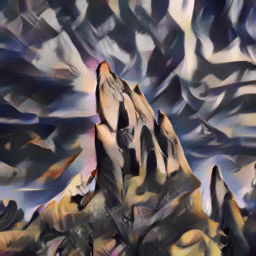}}\hfill
  \subfigure[]  {\includegraphics[width=0.11\linewidth]{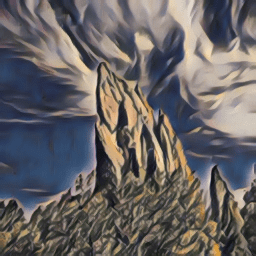}}\hfill\\
  
  \vspace{-22pt}
  
  \subfigure[]  {\includegraphics[width=0.11\linewidth]{latex/suppleResult/content/kitchen.png}}\hfill
  \subfigure[]  {\includegraphics[width=0.11\linewidth]{latex/suppleResult/style/painting.png}}\hfill
  \subfigure[]  {\includegraphics[width=0.11\linewidth]{latex/suppleResult/prior/ours/in25_single_a_reference_25.png}}\hfill
  \subfigure[]  {\includegraphics[width=0.11\linewidth]{latex/suppleResult/prior/gatys/in25_single_a_reference_25_result.png}}\hfill
  \subfigure[]  {\includegraphics[width=0.11\linewidth]{latex/leveraging/strotss/kitchen_painting.png}}\hfill
  \subfigure[]  {\includegraphics[width=0.11\linewidth]{latex/leveraging/SinIR/kitchen_painting.png}}\hfill
  \subfigure[]  {\includegraphics[width=0.11\linewidth]{latex/leveraging/DTP/kitchen.png}}\hfill
  \subfigure[]  {\includegraphics[width=0.11\linewidth]{latex/suppleResult/prior/sanet/19_in25_stylized_19_single_a_reference_25_1000.png}}\hfill
  \subfigure[]  {\includegraphics[width=0.11\linewidth]{latex/suppleResult/prior/attn/19_single_a_reference_25_19_in25_cs.png}}\hfill\\
  \vspace{-22pt}
  
  \subfigure[]  {\includegraphics[width=0.11\linewidth]{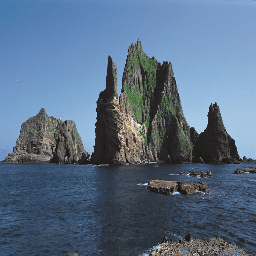}}\hfill
  \subfigure[]  {\includegraphics[width=0.11\linewidth]{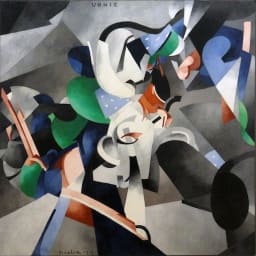}}\hfill
  \subfigure[]  {\includegraphics[width=0.11\linewidth]{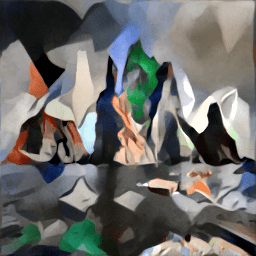}}\hfill
  \subfigure[]  {\includegraphics[width=0.11\linewidth]{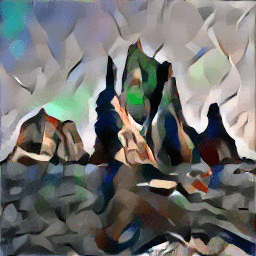}}\hfill
  \subfigure[]  {\includegraphics[width=0.11\linewidth]{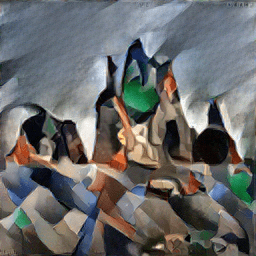}}\hfill
  \subfigure[]  {\includegraphics[width=0.11\linewidth]{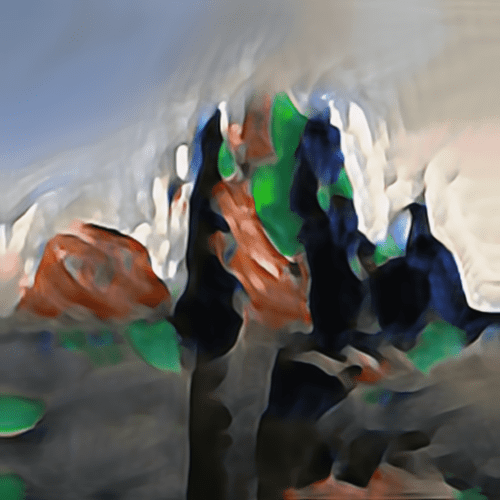}}\hfill
  \subfigure[]  {\includegraphics[width=0.11\linewidth]{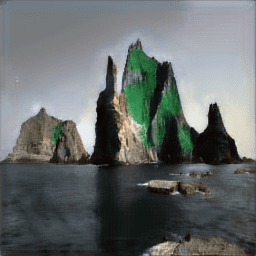}}\hfill
  \subfigure[]  {\includegraphics[width=0.11\linewidth]{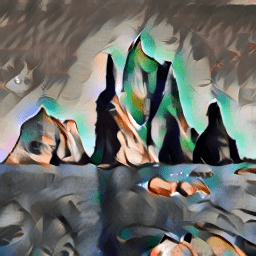}}\hfill
  \subfigure[]  {\includegraphics[width=0.11\linewidth]{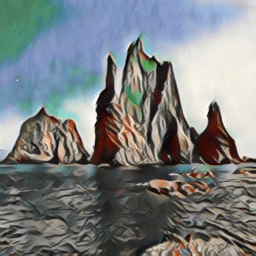}}\hfill\\
  \vspace{-22pt}

  \subfigure[]  {\includegraphics[width=0.11\linewidth]{latex/leveraging/content/bear.jpg}}\hfill
  \subfigure[]  {\includegraphics[width=0.11\linewidth]{latex/leveraging/style/red.jpg}}\hfill
  \subfigure[]  {\includegraphics[width=0.11\linewidth]{latex/leveraging/ours/bear_red_result.png}}\hfill
  \subfigure[]  {\includegraphics[width=0.11\linewidth]{latex/leveraging/nst/bear_red_result.png}}\hfill
  \subfigure[]  {\includegraphics[width=0.11\linewidth]{latex/leveraging/strotss/bear_red.jpg}}\hfill
  \subfigure[]  {\includegraphics[width=0.11\linewidth]{latex/leveraging/SinIR/bear.png}}\hfill
  \subfigure[]  {\includegraphics[width=0.11\linewidth]{latex/leveraging/DTP/bear.png}}\hfill
  \subfigure[]  {\includegraphics[width=0.11\linewidth]{latex/leveraging/SANET/bear_stylized_red.jpg}}\hfill
  \subfigure[]  {\includegraphics[width=0.11\linewidth]{latex/leveraging/AdaAttn/red_bear_cs.png}}\hfill\\
  \vspace{-22pt}
  
  \subfigure[Content]   {\includegraphics[width=0.11\linewidth]{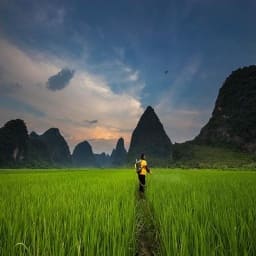}}\hfill
  \subfigure[Style]     {\includegraphics[width=0.11\linewidth]{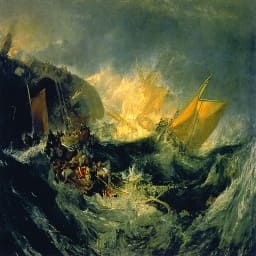}}\hfill
  \subfigure[Ours]      {\includegraphics[width=0.11\linewidth]{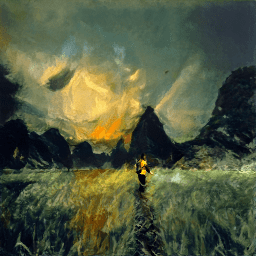}}\hfill 
  \subfigure[Gatys et al.~\cite{gatys2016image}]  {\includegraphics[width=0.11\linewidth]{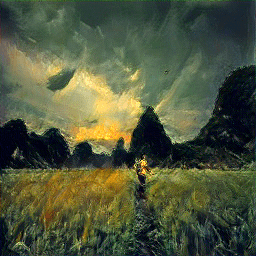}}\hfill
  \subfigure[STROTSS~\cite{kolkin2019style}]   {\includegraphics[width=0.11\linewidth]{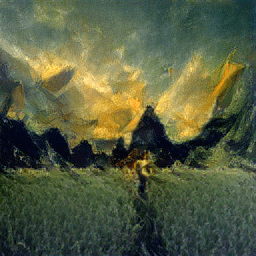}}\hfill
  \subfigure[SinIR~\cite{yoo2021sinir}]     {\includegraphics[width=0.11\linewidth]{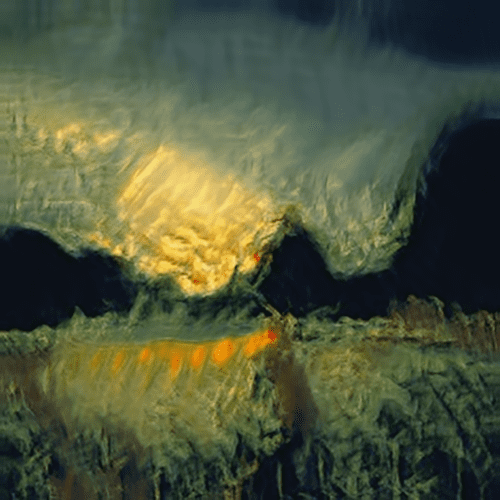}}\hfill
  \subfigure[DTP~\cite{kim2021deep}]       {\includegraphics[width=0.11\linewidth]{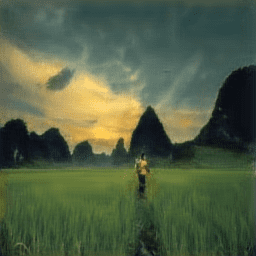}}\hfill
  \subfigure[SANet~\cite{park2019arbitrary}]     {\includegraphics[width=0.11\linewidth]{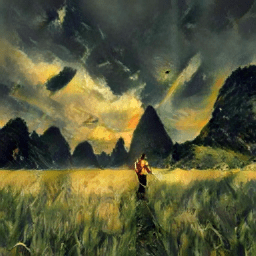}}\hfill
  \subfigure[AdaAttn~\cite{liu2021adaattn}]   {\includegraphics[width=0.11\linewidth]{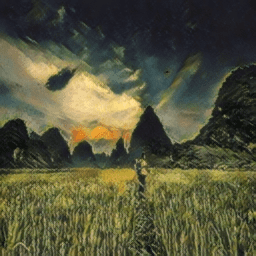}}\hfill\\ 
  
  \vspace{-5pt} 
  
  \caption{\textbf{More qualitative comparisons of existing artistic style transfer methods to our framework.}
  Our model shows more plausible style transfer results in comparison with state-of-the-art methods. Our model demonstrates competitive performance compared to learning-based models.
  }
  \label{fig:qualitative}
\end{figure*}


\begin{figure*}[t!]
\centering
\includegraphics[width=0.8\textwidth]{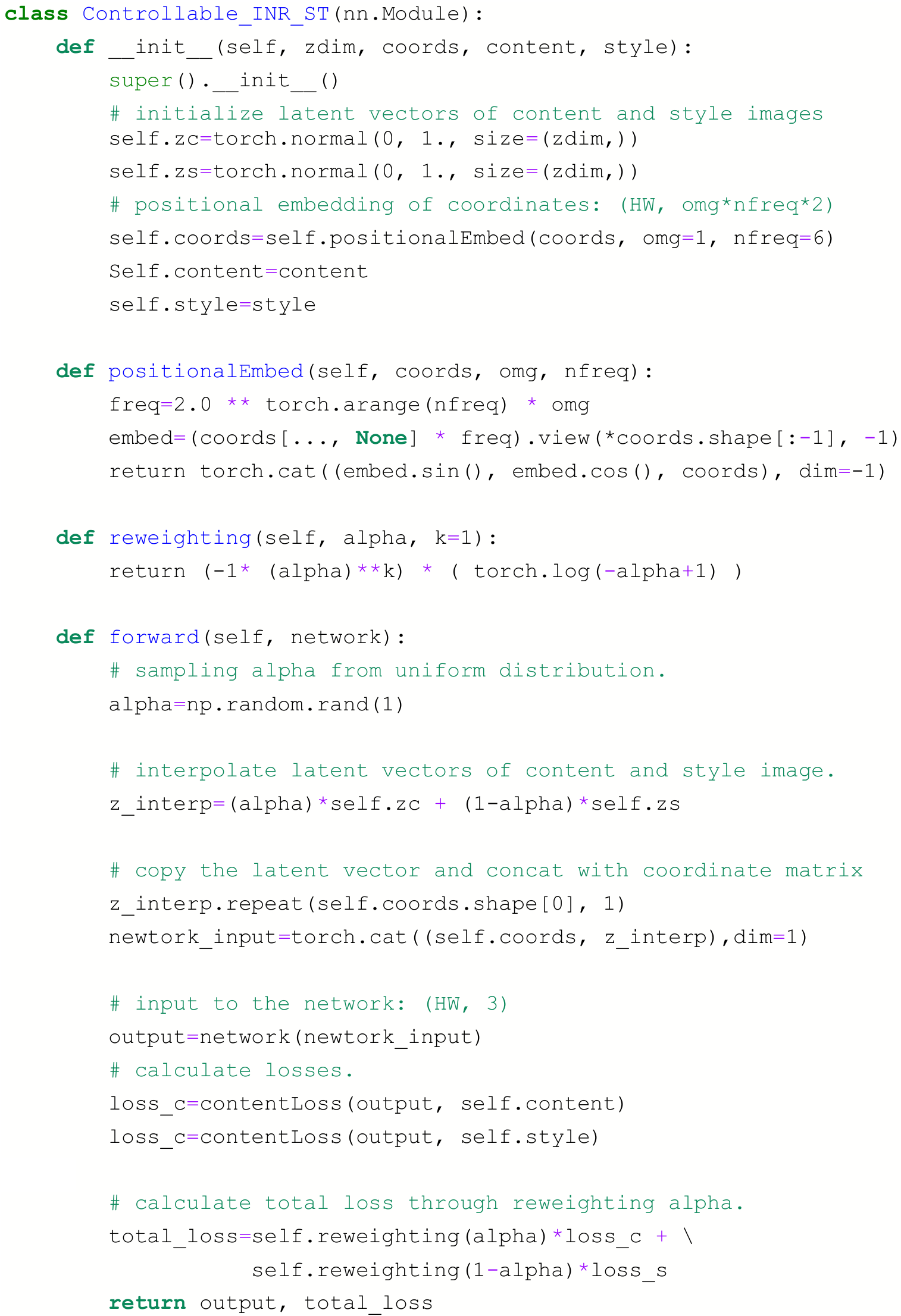}\hfill\\
\caption{\textbf{Pytorch-like pseudo code.}}
\label{fig:pseudo}
\end{figure*}

\clearpage
\newpage



\end{document}